\documentclass[10pt,twocolumn,letterpaper]{article}
\pdfoutput=1

\usepackage{cvpr}

\usepackage{times}
\usepackage{epsfig}
\usepackage{graphicx}
\usepackage{amsmath}
\usepackage{amssymb}
\usepackage{soul}

\usepackage{url}
\usepackage{amsfonts,bm}
\usepackage{epstopdf}
\usepackage{caption}
\usepackage{subcaption}
\usepackage{array}
\usepackage{booktabs,siunitx}
\usepackage{footmisc}

\usepackage[pagebackref=true,breaklinks=true,letterpaper=true,colorlinks]{hyperref}
\usepackage{multibib} %
\newcites{sup}{Supplementary References}

\newcommand{\Ctext}{C_o} %

\usepackage{algorithmic,algorithm}

\hypersetup{
  plainpages=false,
  pdfpagemode=UseNone, %
  colorlinks=true,   %
  linkcolor=blue,    %
  anchorcolor=blue,  %
  citecolor=blue,    %
  filecolor=blue,    %
  pagecolor=blue,    %
  urlcolor=blue,     %
  pdfview=FitH,               %
  pdfstartview=FitH,          %
  pdfpagelayout=SinglePage    %
}

\cvprfinalcopy %

\def\cvprPaperID{1383} %

\ifcvprfinal\fi
\begin{document}

\title{Unsupervised Learning from Narrated Instruction Videos}

\author{
Jean-Baptiste Alayrac\thanks{WILLOW project-team, D\'{e}partement d'Informatique de l'Ecole Normale Sup\'{e}rieure, ENS/INRIA/CNRS UMR 8548, Paris, France.} \ \thanks{SIERRA project-team, D\'epartement d'Informatique de l'Ecole Normale Sup\'{e}rieure, ENS/INRIA/CNRS UMR 8548, Paris, France.}
\and
Piotr Bojanowski\footnotemark[1]
\and
Nishant Agrawal \footnotemark[1] \ \thanks{IIIT Hyderabad} 
\and
Josef Sivic\footnotemark[1]
\and
Ivan Laptev\footnotemark[1] 
\and
Simon Lacoste-Julien\footnotemark[2] 
}

\maketitle

\begin{abstract}
   We address the problem of automatically learning the main steps to complete a certain task, such as changing a car tire, 
from a set of narrated instruction videos. 
The contributions of this paper are three-fold.
First, we develop a new unsupervised learning approach that takes advantage of the complementary nature of the input video and the associated narration.   
The method solves two clustering problems, one in text and one in video, applied one after each other and linked by joint constraints to obtain a single coherent sequence of steps in both modalities.  
Second, we collect and annotate a new challenging dataset of real-world instruction videos from the Internet. The dataset contains about 800,000 frames for five different tasks\footnote{How to : change a car tire, perform CardioPulmonary resuscitation (CPR), jump a car, repot a plant and make coffee} that include complex interactions between people and objects, and are captured in a variety of indoor and outdoor settings.	
Third, we experimentally demonstrate that the proposed method can automatically discover, in an \emph{unsupervised manner}, the main steps to achieve the task and locate the steps in the input videos. 
\end{abstract}

\section{Introduction}
Millions of people watch narrated instruction videos\footnote{Some instruction videos on YouTube have tens of millions of views, 
e.g.~\url{www.youtube.com/watch?v=J4-GRH2nDvw}.} to learn new tasks such as assembling IKEA furniture or changing a flat car tire. 
Many of such tasks have large amounts of videos available on-line. For example, querying for ``how to change a tire'' results in more than 300,000 hits on YouTube.
Most of these videos, however, are made with the intention to teach other people to perform the task and do not provide direct supervisory signal for automatic learning algorithms. Developing unsupervised methods that could learn tasks from myriads of instruction videos on the Internet is therefore a key challenge.
Such automatic cognitive ability would enable constructing virtual assistants and smart robots that learn new skills from the Internet to, for example, help people achieve new tasks in unfamiliar situations.

\begin{figure*}[t]
       \centering
       \includegraphics[width=\linewidth]{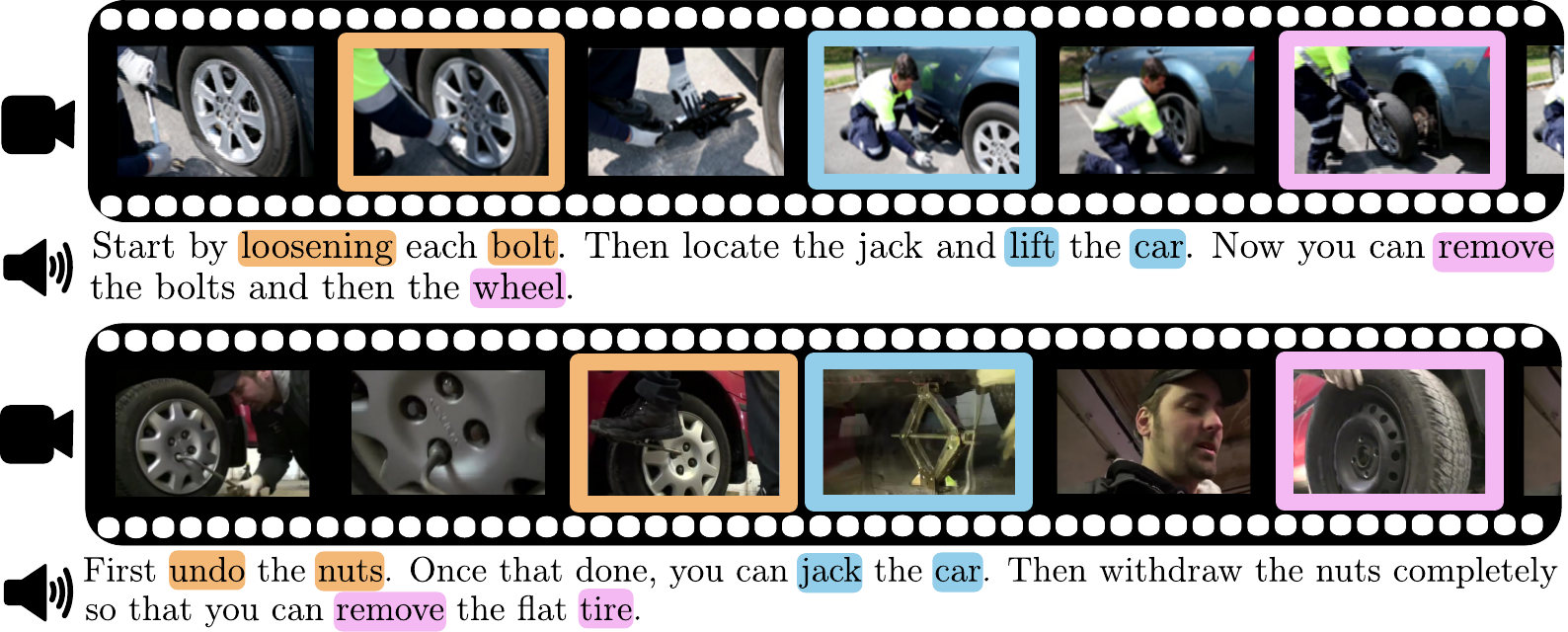}
     \caption{\small Given a set of narrated instruction videos demonstrating a particular task, we wish to automatically discover the main steps to achieve the task and associate each step with its corresponding narration and appearance in each video. Here frames from two videos demonstrating changing the car tire are shown, together with excerpts of the corresponding narrations.  Note the large variations in both the narration and appearance of the different steps highlighted by the same colors in both videos (here only three steps are shown).}
     \vspace{-4mm}
     \label{fig:mainpaper}
 \end{figure*}

In this work, we consider instruction videos and develop a method that learns a sequence of steps, as well as their textual and visual representations, required to achieve a certain task. 
For example, given a set of narrated instruction videos demonstrating how to change a car tire, our method automatically discovers consecutive steps for this task such as {\em loosen the nuts of the wheel}, {\em jack up the car}, {\em remove the spare tire} and so on as illustrated in Figure~\ref{fig:mainpaper}. 
In addition, the method learns the visual and linguistic variability of these steps from natural videos.

Discovering key steps from instruction videos is a highly challenging task.
First, linguistic expressions for the same step can have high variability across videos, for example: ``...Loosen up the wheel nut just a little before you start jacking the car...'' and ``...Start to loosen the lug nuts just enough to make them easy to turn by hand...''.
Second, the visual appearance of each step varies greatly between videos as the people and objects are different, the action is captured from a different viewpoint, and the way people perform actions also vary.
Finally, there is also a variability of the overall structure of the sequence of steps achieving the task. For example, 
some videos may omit some steps or change slightly their order.

To address these challenges, in this paper we develop an unsupervised learning approach that takes advantage of the complementarity of the 
visual signal in the video and the corresponding natural language narration to resolve their ambiguities. 
We assume that the same ordered sequence of steps (also called script in the NLP literature~\cite{Regneri10learning}) is common to all input videos of the same task, but the actual sequence and the individual steps are unknown and are learnt directly from data. 
This is in contrast to other existing methods for modeling instruction videos~\cite{Malmaud15what} that assume a script (recipe) is known and fixed in advance. 
We address the problem by first performing temporal clustering of text followed by clustering in video, where the two clustering tasks are linked by joint constraints. 
The complementary nature of the two clustering problems helps to resolve ambiguities in the two individual modalities. For example, two video segments with very different appearance but depicting the same step can be grouped together  because they are narrated in a similar language.
Conversely, two video segments described with very different expressions, for example, ``jack up the car'' and ``raise the vehicle'' can be identified as belonging to the same instruction step because they have similar visual appearance.
The output of our method is the script listing the discovered steps of the task as well as the temporal location of each step in the input videos. 
We validate our method on a new dataset of instruction videos composed of five different tasks with a total of 150 videos and about 800,000 frames.

\section{Related work}

This work relates to unsupervised and weakly-supervised learning methods in computer vision and natural language processing.
Particularly related to ours is the work on learning script-like knowledge from natural language descriptions~\cite{Chambers08,Frermann14,Regneri10learning}. These methods aim to discover typical events (steps) and their order for particular scenarios (tasks)\footnote{We here assign the same meaning to terms ``event'' and ``step'' as well as to terms ``script'' and ``task''.} such as ``cooking scrambled egg'', ``taking a bus'' or ``making coffee''. While~\cite{Chambers08} uses large-scale news copora,~\cite{Regneri10learning} argues that many events are implicit and are not described in such general-purpose text data. Instead,~\cite{Frermann14,Regneri10learning} use event sequence descriptions collected for particular scenarios. Differently to this work, we learn sequences of events from narrated instruction videos on the Internet. Such data contains detailed event descriptions but is not structured and contains more noise compared to the input of~\cite{Frermann14,Regneri10learning}. 

Interpretation of narrated instruction videos has been recently addressed in~\cite{Malmaud15what}. While this work analyses cooking videos at a great scale, it relies on readily-available recipes which may not be available for more general scenarios. Differently from~\cite{Malmaud15what}, we here aim to learn the steps of instruction videos using a discriminative clustering approach.
A similar task to ours is addressed in~\cite{Naim15discriminative} using latent variable structured perceptron algorithm to align nouns in instruction sentences with objects touched by hands in instruction videos.
However, similarly to~\cite{Malmaud15what}, \cite{Naim15discriminative} uses laboratory experimental protocols as textual input, whereas here we consider a weaker signal in the form of the real transcribed narration of the video.

In computer vision, unsupervised action recognition has been explored in simple videos~\cite{Niebles08}. 
More recently, weakly supervised learning of actions in video using video scripts or event order has been addressed in~\cite{Bojanowski13finding,Bojanowski14weakly,Bojanowski15weakly,Duchenne2009automatic,Laptev08a}. 
Particularly related to ours is the work~\cite{Bojanowski14weakly} which explores the known order of events to localize and learn actions in training data. 
While~\cite{Bojanowski14weakly} uses manually annotated sequences of events, we here discover the sequences of main events by clustering  transcribed narrations of the videos. Related is also the work of
\cite{Bojanowski15weakly} that aligns natural text descriptions to video but in contrast to our approach does not discover automatically the common sequence of main steps. 
Methods in~\cite{Niebles10a,Raptis13} learn in an unsupervised manner the temporal structure of actions from video but do not discover textual expressions for actions as we do in this work.
The recent concurrent work~\cite{Sener15unsupervised} is addressing, independently of our work, a similar problem but with a different approach based on a probabilistic generative model and considering a different set of tasks mainly focussed on cooking activities.

Our work is also related to video summarization and in particular to the recent work on category-specific video summarization \cite{Potapov14category,Sun14ranking}. While summarization is a subjective task, we here aim to extract the key steps required to achieve a concrete task that  consistently appear in the same sequence in the input set of videos. In addition, unlike video summarization~\cite{Potapov14category,Sun14ranking} we jointly exploit visual and linguistic modalities in our approach.

\vspace{3mm}
\section{New dataset of instruction videos}
\label{sec:dataset}
We have collected a dataset of narrated instruction videos for five tasks: \textit{Making a coffee}, \textit{Changing car tire}, \textit{Performing cardiopulmonary resuscitation (CPR)}, \textit{Jumping a car} and \textit{Repotting a plant}. 
The videos were obtained by searching YouTube with relevant keywords. 
The five tasks were chosen so that they have a large number of available videos with English transcripts while trying to cover a wide range of activities that include complex interactions of people with objects and other people.
For each task, we took the top 30 videos with English ASR returned by YouTube.
We also quickly verified that each video contains a person actually performing the task (as opposed to just talking about it).
The result is a total of 150 videos, 30 videos for each task.
The average length of our videos is about 4,000 frames (or 2 minutes) and the entire dataset contains about 800,000 frames.

The selected videos have English transcripts obtained from YouTube's automatic speech recognition (ASR) system.
To remove the dependence of results on errors of the particular ASR method, we have manually corrected misspellings and punctuations in the output transcriptions.
We believe this step will soon become obsolete given rapid improvements of ASR methods. 
As we do not modify the content of the spoken language in videos, the transcribed verbal instructions still represent an extremely challenging example of natural language with large variability in the used expressions and terminology.
Each word of the transcript is associated with a time interval in the video (usually less than 5 seconds) obtained from the closed caption timings.

For the purpose of evaluation, we have manually annotated the temporal location in each video of the main steps necessary to achieve the given task.
For all tasks, we have defined the ordered sequence of ground truth steps before running our algorithm.
The choice of steps was made by an agreement of 2-3 annotators who have watched the input videos and verified the steps on instruction video websites such as \url{http://www.howdini.com}.
 While some steps can be occasionally left out in some videos or the ordering slightly modified, overall we have observed a good consistency 
 in the given sequence of instructions among the input videos.
We measured that only 6\% of the step annotations did not fit the global order, while a step was missing from the video 27\% of the time.\footnote{We describe these measurements in more details in the supplementary material given in Appendix~\ref{subsec:score_dataset}.}
We hypothesize that this could be attributed to the fact that all videos are made with the same goal of giving other humans clear, concise and comprehensible verbal and visual instructions on how to achieve the given task.  
Given the list of steps for each task, we have manually annotated each time interval in each input video to one of the ground truth steps (or no step).
The actions of the individual steps are typically separated by hundreds of frames where the narrator transitions between the steps or explains verbally what is going to happen.  
Furthermore, some steps could be missing in some videos, or could be present but not described in the narration.  Finally, the temporal alignment between the narration and the actual actions in video is only coarse as the action is often described before it is performed.

\begin{figure*}[t]
\includegraphics[width=1.02\linewidth]{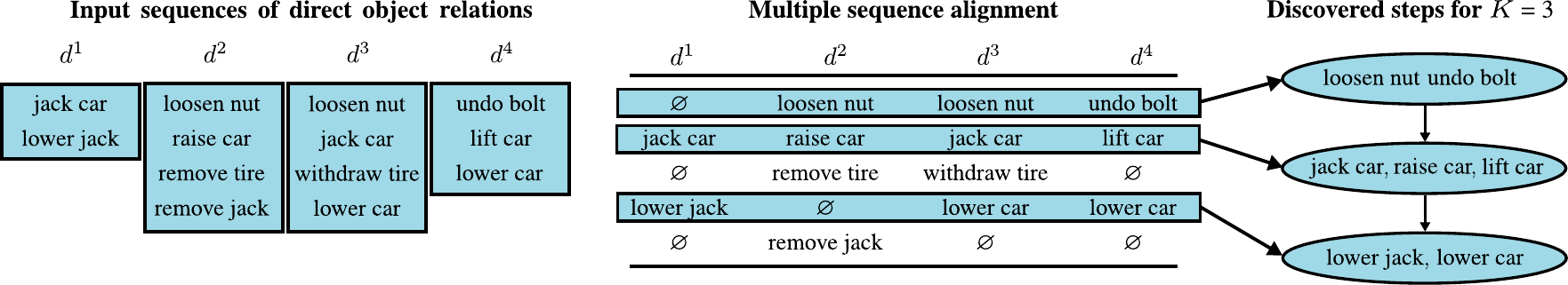}
   \caption{\small {\bf Clustering transcribed verbal instructions.} {\bf Left:} The input raw text for each video is converted into a sequence of direct object relations. Here, an illustration of four sequences from four different videos is shown.  {\bf Middle:} Multiple sequence alignment is used to align all sequences together. Note that different direct object relations are aligned together as long as they have the same sense, e.g. ``loosen nut" and ``undo bolt".  {\bf Right:} The main instruction steps are extracted as the $K=3$ most common steps in all the sequences. \vspace{-.3cm}
   }
   \label{tab:MSAillustration}
   \vspace*{-0.1cm}
\end{figure*}

\section{Modelling narrated instruction videos}
\label{subsec:model_var}

We are given a set of $N$ instruction videos all depicting the same task (such as ``changing a tire'').
The $n$-th input video is composed of a video stream of $T_n$ segments of frames $(x^n_t)_{t=1}^{T_n}$ and an audio stream containing a detailed verbal description of the depicted task. 
We suppose that the audio description was transcribed to raw text and then processed to a sequence of $S_n$ text tokens $(d^n_s)_{s=1}^{S_n}$. 
Given this data, we want to automatically recover the sequence of $K$ main steps that compose the given task and locate each step within each input video and text transcription. 

 We formulate the problem as two clustering tasks, one in text and one in video, applied one after each other and linked by joint constraints linking the two modalities.
  This two-stage approach is based on the intuition that the variation in natural language describing each task is easier to capture than the visual variability of the input videos. 
 In the first stage, we cluster the text transcripts into a sequence of $K$ main steps to complete the given task.  Empirically, we have found (see results in Sec.~\ref{sec:step_discovery}) that it is possible to discover the sequence of the $K$ main steps for each task with high precision. However, the text itself gives only a poor localization of each step in each video. 
 Therefore, in the second stage we accurately localize each step in each video by clustering the input videos using the sequence of $K$ steps extracted from text as constraints on the video clustering. 
To achieve this, we use two types of constraints between video and text.  First, we assume that both the video and the text narration follow the same sequence of steps. This results in a global ordering constraint on the recovered clustering. Second, we assume that people perform the action approximately at the same time that they talk about it. This constraint temporally links the recovered clusters in text and video. %
The important outcome of the video clustering stage is that the $K$ extracted steps get propagated by visual similarity to videos where the text descriptions are missing or ambiguous. 

We first describe the text clustering in Sec.~\ref{subsec:model_text} and then introduce the video clustering with constraints in Sec.~\ref{sec:model-video}.

\subsection{Clustering transcribed verbal instructions}
\label{subsec:model_text}
\label{sec:text}

The goal here is to cluster the transcribed verbal descriptions of each video into a sequence of \emph{main steps} necessary to achieve the task.  
This stage is important as the resulting clusters will be used as constraints for jointly learning and localizing the main steps in video. 
We assume that the important steps are common to many of the transcripts and that the sequence of steps is (roughly) preserved in all transcripts.  
Hence, following~\cite{Regneri10learning}, we formulate the problem of clustering the input transcripts as a multiple sequence alignment problem.
However, in contrast to~\cite{Regneri10learning} who cluster manually provided descriptions of each step, we wish to cluster transcribed verbal instructions.
Hence our main challenge is to deal with the variability in spoken natural language.  To overcome this challenge, we take advantage of the fact that completing a certain task usually involves interactions with objects or people and hence we can extract a more structured representation from the input text stream.      

More specifically, we represent the textual data as a sequence of {\em direct object relations}.
A direct object relation~$d$  is a pair composed of a verb and its direct object complement, such as ``remove tire".
Such a direct object relation can be extracted from the dependency parser of the input transcribed narration~\cite{Marneffe06generating}.
We denote the set of all different direct object relations extracted from all narrations as $\mathcal{D}$,
with cardinality~$D$. 
For the $n$-th video, we thus represent the text signal as a sequence of direct object relation tokens: $d^n = (d_1^n, \dots, d_{S_n}^n)$, where the length $S_n$ of the sequence varies from one video clip to another. 
This step is key to the success of our method as it allows us to convert the problem of clustering raw transcribed text into an easier problem of clustering sequences of direct object relations.    
The goal is now to extract from the narrations the most common sequence of  $K$ main steps to achieve the given task. 
To achieve this, we first find a globally consistent alignment of the direct object relations that compose all text sequences by solving a multiple sequence alignment problem.
Second, we pick from this alignment the $K$ most globally consistent clusters across videos. %

\textbf{Multiple sequence alignment model.} We formulate the first stage of finding the common alignment between the input sequences of direct object relations as a multiple sequence alignment problem with the \emph{sum-of-pairs score}~\cite{wang1994msaNPhard}. 
In details, a global alignment can be defined by re-mapping each input sequence $d^n$ of tokens to a global common template of $L$ slots, for $L$ large enough. We let $(\phi(d^n))_{1\leq l \leq L}$ represent
the (increasing) re-mapping for sequence $d^n$ at the new locations indexed by $l$:  $\phi(d^n)_l$ represents the direct object relation put at location $l$, with $\phi(d^n)_l = \varnothing$ if a slot is left empty (denoting the insertion of a gap in the original sequence of tokens). See the middle of Figure~\ref{tab:MSAillustration} for an example of re-mapping.
The goal is then to find a global alignment that minimizes the following sum-of-pairs cost function:
\vspace{-3mm}
\begin{equation}
\sum_{(n,m)} \sum_{l=1}^L c(\phi(d^n)_l,\phi(d^{m})_{l}),
\label{eq:msa_cost}
\end{equation}
where $c(d_1,d_2)$ denotes the cost of aligning the direct object relations $d_1$ and $d_2$ at the same common slot~$l$ in the global template. The above cost thus denotes the sum of all pairwise alignments of the individual sequences (the outer sum), where the quality of each alignment is measured by summing the cost $c$ of matches of individual direct object relations mapped into the common template sequence. We use a negative cost when $d_1$ and $d_2$ 
are similar according to the distance in the WordNet tree~\cite{Fellbaum98Wordnet,Miller95Wordnet} of their verb and direct object constituents, and positive if they are dissimilar (details are given in 
Sec.~\ref{sec:experiments}).
As the verbal narrations can talk about many other things than the main steps of a task,
we set $c(d,d')=0$ if either $d$ or $d'$ is $\varnothing$.
An illustration of clustering the transcribed verbal instructions into a sequence of $K$ steps is shown in Figure~\ref{tab:MSAillustration}.

\textbf{Optimization using Frank-Wolfe.} 
Optimizing the cost~\eqref{eq:msa_cost} is NP-hard~\cite{wang1994msaNPhard} 
because of the combinatorial nature of the problem. The standard
solution from computational biology is to apply a heuristic algorithm that proceeds by incremental pairwise alignment using dynamic programming~\cite{Lee01poa}.
In contrast, we show in Appendix~\ref{subsec:details_msa} that the multiple sequence alignment problem given by~\eqref{eq:msa_cost} can be reformulated as an integer quadratic program with combinatorial constraints, for which the Frank-Wolfe optimization algorithm has been used recently with increasing success~\cite{Bojanowski14weakly,Jaggi2013,Joulin14efficient,Lacoste15GlobalLinearFW}. 
Interestingly, we have observed empirically (see Appendix~\ref{subsec:comparison_msa}) that
the Frank-Wolfe algorithm was giving better solutions (in terms of objective~\eqref{eq:msa_cost}) than the state-of-the-art heuristic procedures for this task~\cite{Higgins88clustal,Lee01poa}. Our Frank-Wolfe based solvers also offer us greater flexibility in defining the alignment cost and scale better with the length of input sequences and the vocabulary of direct object relations. 

\textbf{Extracting the main steps.}
After a global alignment is obtained, we sort the global template $l$
by the number of direct object relations aligned to each slot. Given
$K$ as input, the top $K$ slots give the main instruction steps for
the task, unless there are multiple steps with the same support, which
go beyond $K$. In this case, we pick the next smaller number below $K$
which excludes these ties, allowing the choice of an \emph{adaptive}
number of main instruction steps when there is not enough saliency for
the last steps.  This strategy essentially selects $k\leq K$ salient
steps, while refusing to make a choice among steps with equal support
that would increase the total number of steps beyond $K$.  As we will
see in our results in Sec.~\ref{subsec:exp0}, our algorithm
sometimes returns a much smaller number than $K$ for the main
instruction steps, giving more robustness to the exact choice of
parameter $K$.

\textbf{Encoding of the output.}
We post-process the output of multiple sequence alignment into an assignment matrix $R_n \in \{0, 1\}^{S_n \times K}$ for each input video $n$, 
where $(R_n)_{sk} = 1$ means that the direct object token $d^n_s$ has been assigned to step $k$. 
If a direct object has not been assigned to any step, the corresponding row of the matrix $R_n$ will be zero.

\subsection{Discriminative clustering of videos under text constraints}
\label{sec:model-video}

Given the output of the text clustering that identified the important $K$ steps forming a task,
we now want to find their temporal location in the video signal. We formalize this problem
as looking for an assignment matrix $Z_n\in \{0, 1\}^{T_n \times K}$ for each input video $n$,
where $(Z_n)_{tk} = 1$ indicates the visual presence of step $k$ at time interval~$t$ in video~$n$, and $T_n$ is the length of video~$n$.
Similarly to $R_n$, we allow the possibility that a whole row of $Z_n$ is zero, indicating that no step 
is visually present for the corresponding time interval.

We propose to tackle this problem using a discriminative clustering approach with global
ordering constraints, as was successfully used in the past
for the temporal localization of actions in videos~\cite{Bojanowski14weakly}, 
but with additional \emph{weak temporal constraints}.
In contrast to~\cite{Bojanowski14weakly} where the order of actions was manually
given for each video, our multiple sequence alignment approach automatically
discovers the main steps. More importantly, we also use the \emph{text caption timing}
to provide a fine-grained weak temporal supervision for the visual 
appearance of steps, which is described next.

\setlength{\tabcolsep}{2pt}
\begin{table*}[t]\centering
\resizebox{\textwidth}{!}{
    \footnotesize
    \begin{tabular}{lr >{\centering\hspace{0.5pt}}m{0cm} lr >{\centering\hspace{0.5pt}}m{0cm} lr >{\centering\hspace{0.5pt}}m{0cm} lr >{\centering\hspace{0.5pt}}m{0cm} lr}
    \toprule
         \multicolumn{2}{c}{Changing a tire}  & \phantom{abc} & \multicolumn{2}{c}{Performing CPR} & \phantom{abc} & \multicolumn{2}{c}{Repot a plant} & \phantom{abc} & \multicolumn{2}{c}{Make coffee} & \phantom{abc} & \multicolumn{2}{c}{Jump car} \\
    
\cmidrule{1-2} \cmidrule{4-5} \cmidrule{7-8} \cmidrule{10-11} \cmidrule{13-14} 
GT (\textbf{11}) & $K\leq 10$ && GT (\textbf{7}) & $K\leq 10$  && GT (\textbf{7}) & $K\leq 10$  && GT (\textbf{10)} & $K\leq 10$ && GT (\textbf{12}) & $K\leq 10$  \\
\cmidrule{1-2} \cmidrule{4-5} \cmidrule{7-8} \cmidrule{10-11} \cmidrule{13-14} 
\textit{get tools out}  & \textbf{get tire}     &&  \textit{open airway}   & \textbf{open airway}      &&  \textit{take plant}   & \textbf{remove plant}    && \textit{add coffee}    & \textbf{put coffee}    && \textit{connect red A} & \textbf{connect cable}    \\

\textit{start loose}  & \textbf{loosen nut}     &&  \textit{check pulse}   & \textbf{put hand}     &&  \textit{put soil}  & \textbf{use soil}    &&   & \textbf{fill chamber}      &&   & \textbf{charge battery}    \\

\textit{}          & \textbf{put jack}     &&        & \textbf{tilt head}     &&  \textit{loosen roots}    & \textbf{loosen soil}   && \textit{fill water}     & \textbf{fill water}   && \textit{connect red B}  & \textbf{connect end}    \\

\textit{jack car}  & \textbf{jack car}     &&        & \textbf{lift chin}     &&  \textit{place plant}    &\textbf{place plant}   && \textit{screw filter}  & \textbf{put filter}   && \textit{start car A}  & \textbf{start car}    \\

\textit{unscrew wheel}  & \textbf{remove nut}     &&  \textit{give breath}   & \textbf{give breath}     &&  \textit{add top}   & \textbf{add soil}    && \textit{}     & \textbf{see steam } &&  \textit{remove cable A}  &\textbf{remove cable}    \\

\textit{remove wheel}  & \textbf{take wheel}     &&  \textit{do compressions}   & \textbf{do compression}    &&  \textit{water plant}   & \textbf{water plant}    && \textit{put stove}     & \textbf{take minutes }                  &&  \textit{remove cable B}  &\textbf{disconnect cable}   \\

\textit{put wheel}  & \textbf{take tire}     &&  \textit{}   & \textbf{open airway}      &&      &    && \textit{}    & \textbf{make coffee}   &&    &  \\

\textit{screw wheel}  & \textbf{put nut}     &&  \textit{}   & \textbf{start compression}     &&      &     &&   \textit{see coffee}  &  \textbf{see coffee}   &&    &     \\

\textit{lower car}  & \textbf{lower jack}     &&  \textit{}   & \textbf{do compression}     &&      &    &&    \textit{pour coffee}  &  \textbf{make cup}    &&    &     \\

\textit{tight wheel}  & \textbf{tighten nut}     &&  \textit{}   & \textbf{give breath}      &&      &     &&      &    &&   &     \\  
\midrule
Precision  & 0.9    &&  Precision &   0.4   &&  Precision &   1 && Precision &   0.67  &&  Precision &   0.83   \\
Recall     & 0.9    &&  Recall    &   0.57  &&  Recall &   0.86  && Recall &   0.6 && Recall &   0.42   \\

        \bottomrule
    \end{tabular}
    
}

    \vspace{-2mm}

    \caption{\small 
        Automatically recovered sequences of steps for the five tasks.
        Each recovered step is represented by one of the aligned direct object relations  (shown in bold). 
        Note that most of the recovered steps correspond well to the ground truth steps (shown in italic).
        The results are shown for the maximum number of discovered steps $K$ set to $10$. Note how our method automatically selects less than 10 steps in some cases. These are the automatically chosen $k\leq K$ steps that are the most salient in the aligned narrations as described in Sec.~\ref{subsec:model_text}.  For {\em CPR}, our method recovers fine-grained steps e.g.~{\em tilt head}, {\em lift chin}, which are not included in the main ground truth steps, but nevertheless could be helpful in some situations, as well as repetitions that were not annotated but were indeed present.
    }
    \label{exp:MSARes}

    \vspace{-4mm}

\end{table*}

\setlength{\tabcolsep}{6pt}

\textbf{Temporal weak supervision from text.} 
From the output of the multiple sequence alignment (encoded in the matrix $R_n \in \{0,1\}^{S_n \times K}$),
each direct object token $d_s^n$ has been assigned to one of the possible $K$ steps,
or to no step at all. 
We use the tokens that have been assigned to a step as 
a constraint on the visual appearance of the same step in the video (using
the assumption that people do what they say approximately when they say it).
We encode the closed caption timing alignment by a binary matrix 
$A_n \in \{0,1\}^{S_n \times T_n}$ for each video, 
where $(A_n)_{st}$ is $1$ if the $s$-th direct object is mentioned in a closed caption 
that overlaps with the time interval $t$ in video.
Note that this alignment is only approximate as people 
usually do not perform the action exactly at the same time that they talk about it, 
but instead with a varying delay.
Second, the alignment is noisy as people typically perform the action only 
once, but often talk about it multiple times (e.g. in a summary at the beginning of the video).
We address these issues by the following two \emph{weak supervision} constraints.
First, we consider a larger set of possible time intervals $[t-\Delta_b, t+\Delta_a]$ in the matrix $A$ rather than the exact time interval $t$ given by the timing of the closed caption.
$\Delta_b$ and $\Delta_a$ are global parameters fixed either qualitatively, or by cross-validation
if labeled data is provided. 
Second, we put as a constraint that the action happens at least once in the set of all possible video time intervals where the action is mentioned in the transcript (rather than every time it is mentioned).
These constraints can be encoded as the following linear inequality constraint on $Z_n$: $A_n Z_n \geq R_n$
(see Appendix~\ref{subsec:fw_dp} for the detailed derivation). 

\textbf{Ordering constraint.} In addition, we also enforce that the temporal order of the steps appearing
visually is consistent with the discovered script from the text, encoding
our assumption that there is a common ordered script for the task across videos.
We encode these sequence constraints on $Z_n$ in a similar manner to~\cite{Bojanowski15weakly},
which was shown to work better than the encoding used in~\cite{Bojanowski14weakly}.
In particular, we only predict the \emph{most salient} time interval in the video that describes a given step.
This means that a particular step is assigned to \emph{exactly one} time interval in each video.
We denote by $\mathcal{Z}_n$ this sequence ordering constraint set.

\textbf{Discriminative clustering.}
The main motivation behind discriminative clustering is to find a \emph{clustering} of the data that can be easily recovered by a \emph{linear classifier} through the minimization of an appropriate \emph{cost function} over the assignment matrix $Z_n$.
The approach introduced in \cite{Bach07diffrac} allows to easily add prior information on the expected clustering.
Such priors have been recently introduced in the context of aligning video and text~\cite{Bojanowski14weakly, Bojanowski15weakly} in the form of ordering constraints over the latent label variables.
Here we use a similar approach to cluster the $N$ input video streams $(x_t)$ into a sequence of $K$ steps, as follows.
We represent each time interval by a $d$-dimensional feature vector. 
The feature vectors for the $n$-th video are stacked in a $T_n\times d$ design matrix denoted by $X_n$. 
We denote by $X$ the $T \times d$ matrix obtained by the concatenation of all $X_n$ matrices
(and similarly, by $Z$, $R$ and $A$ the appropriate concatenation of the $Z_n$, $R_n$ and $A_n$ matrices over $n$). 
In order to obtain the temporal localization into $K$ steps, we learn a linear classifier represented by a $d \times K$ matrix denoted by $W$.
This model is shared among all videos. 

The target assignment~$\hat{Z}$ is found by minimizing the clustering cost function $h$ under 
both the consistent script ordering constraints $\mathcal{Z}$ and
our weak supervision constraints:
\begin{equation}
\underset{Z}{\text{minimize}} \quad h(Z) \quad  \text{ s.t. }  \underbrace{Z \in \mathcal{Z}}_{\text{ordered script}}, \quad
\underbrace{AZ \geq R}_{\substack{\text{weak textual}\\ \text{constraints}} }.
\label{eq:videocost}
\end{equation}
The clustering cost $h(Z)$ is given as in DIFFRAC~\cite{Bach07diffrac} as:
\begin{align}
    h(Z) = \min_{W \in \mathbb{R}^{K \times d}} \ \underbrace{\frac{1}{2T} \|Z - X W\|_F^2}_\text{Discriminative loss on data} + \underbrace{\frac{\lambda}{2} \|W\|_F^2}_\text{Regularizer}. 
    \label{eq:discriminative}
\end{align}
The first term in~(\ref{eq:discriminative}) is the discriminative loss on the data that measures how easy the input data $X$ is separable by the linear classifier $W$ when the target classes are given by the assignments $Z$.  
For the squared loss considered in eq.~(\ref{eq:discriminative}), the optimal weights $W^*$ minimizing~(\ref{eq:discriminative}) can be found in closed form, 
which significantly simplifies the computation. 
However, to solve~\eqref{eq:videocost}, we need to optimize over assignment matrices $Z$ that encode sequences of events and incorporate constraints given by clusters obtained from transcribed textual narrations (Sec.~\ref{sec:text}). 
This is again done by using the Frank-Wolfe algorithm, which allows the use of \emph{efficient dynamic programs} 
to handle the combinatorial constraints on $Z$.
More details are given in Appendix~\ref{sec:details_diffrac}.
\begin{figure*}[t!]
    \centering
    \begin{subfigure}[t]{0.19\linewidth}
        \centering
        \includegraphics[width=\linewidth]{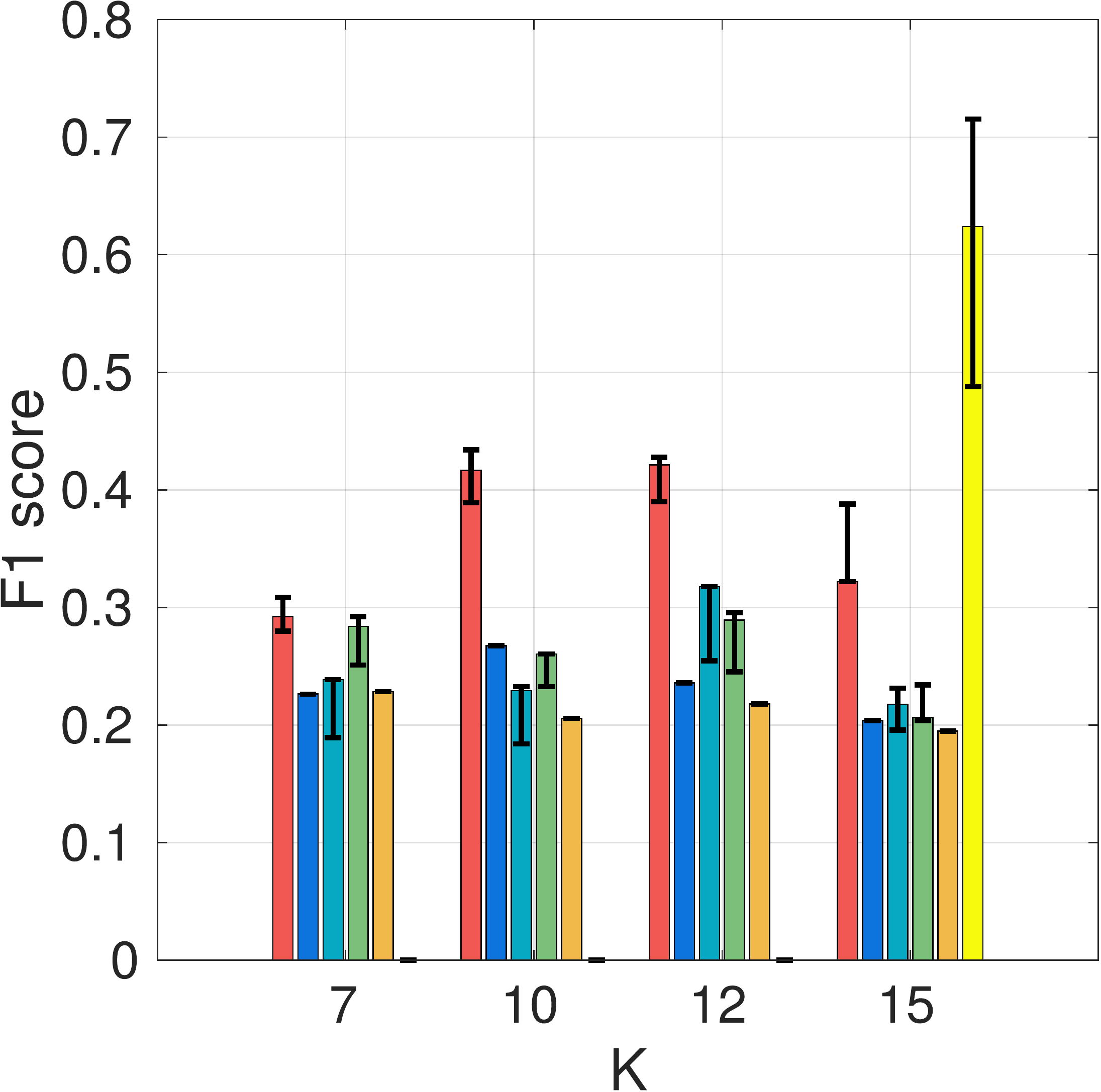}
        \caption{Change tire (\textbf{11})}
    \end{subfigure}%
    ~ 
    \begin{subfigure}[t]{0.19\linewidth}
        \centering
        \includegraphics[width=\linewidth]{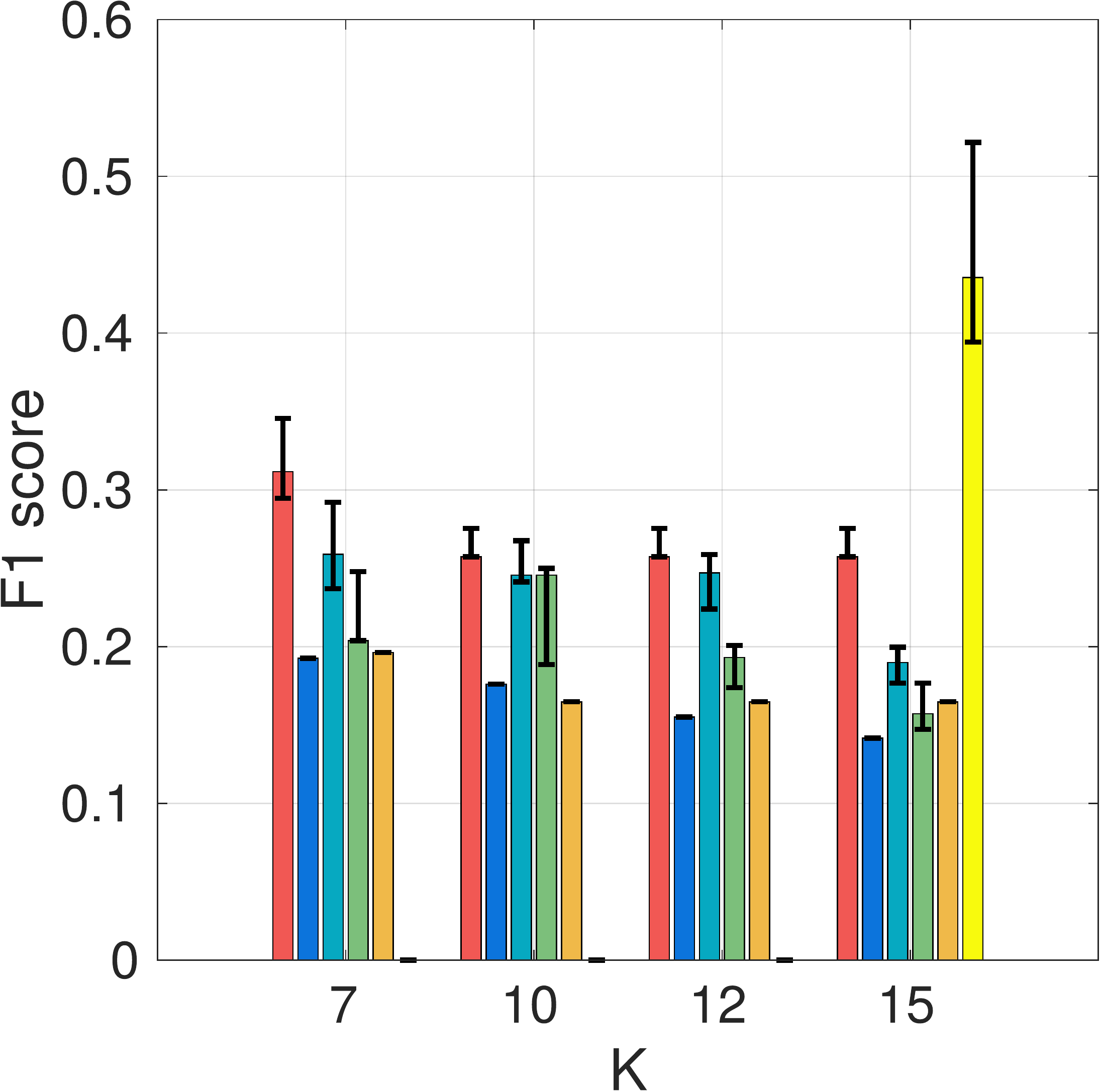}
        \caption{Perform CPR (\textbf{7})}
    \end{subfigure}%
	~
	\begin{subfigure}[t]{0.19\linewidth}
        \centering
        \includegraphics[width=\linewidth]{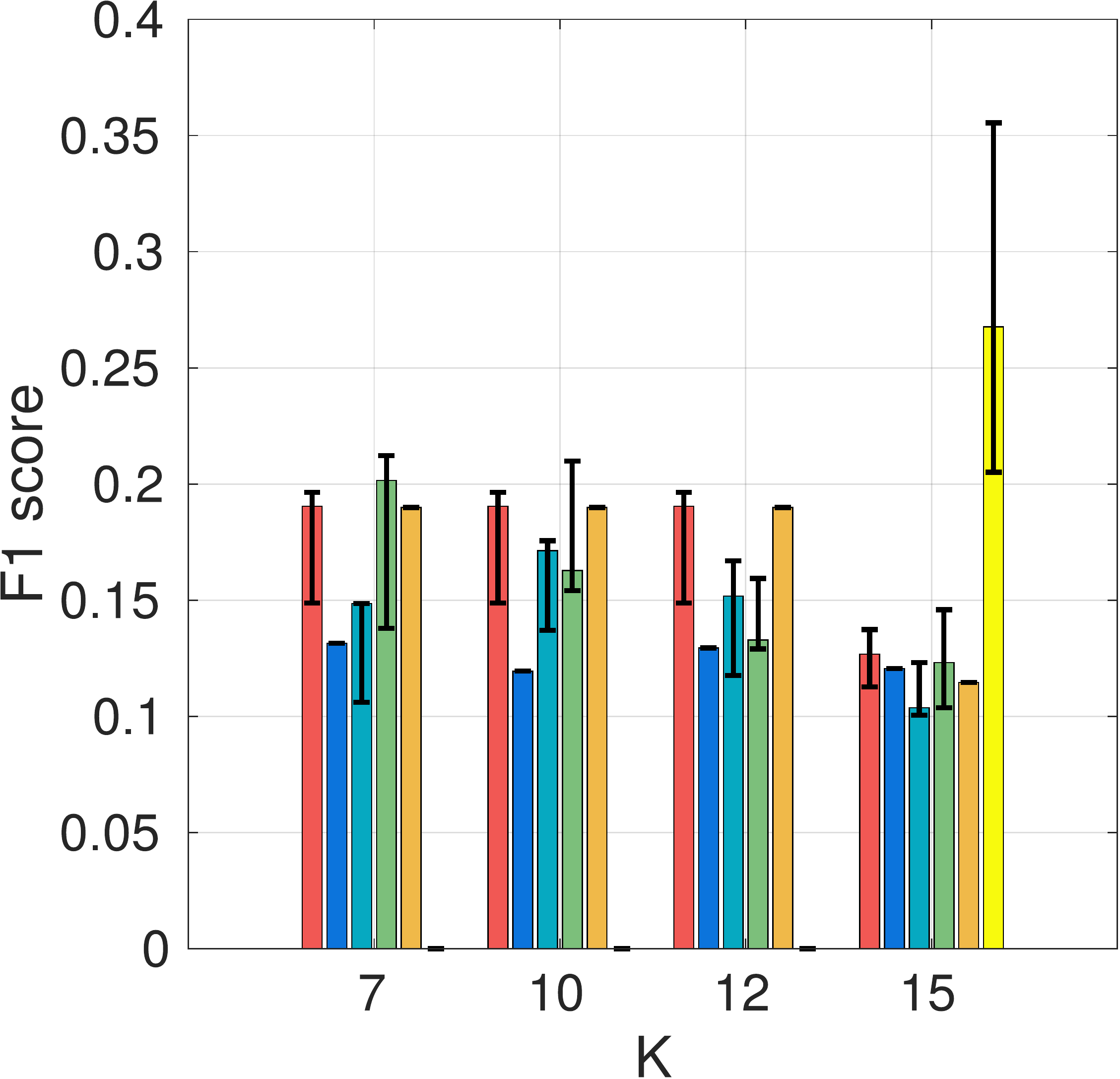}
        \caption{Repot plant (\textbf{7})}
    \end{subfigure}%
    ~
    \begin{subfigure}[t]{0.19\linewidth}
        \centering
        \includegraphics[width=\linewidth]{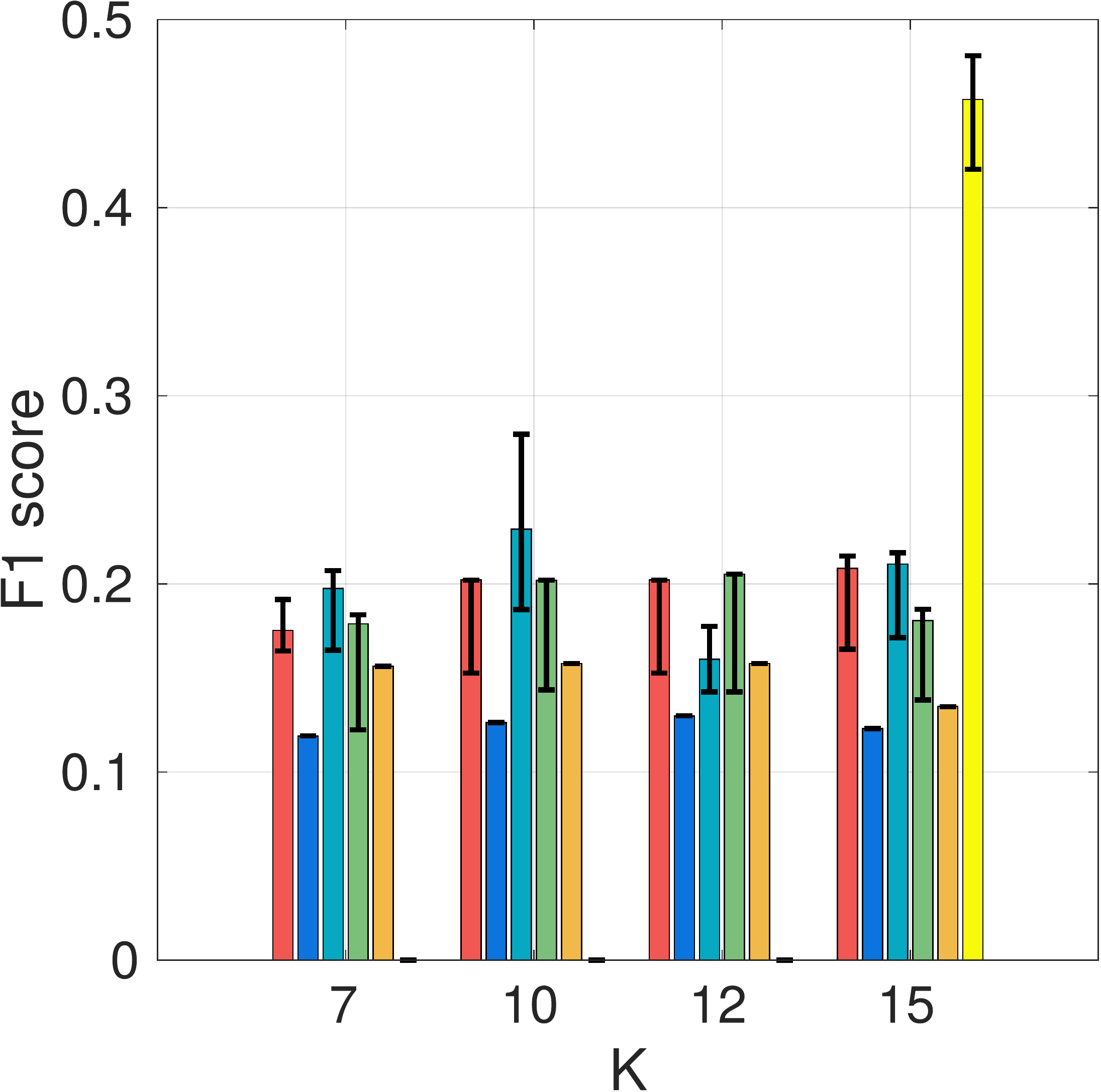}
        \caption{Make coffee (\textbf{10})}
    \end{subfigure}%
    ~
    \begin{subfigure}[t]{0.19\linewidth}
        \centering
        \includegraphics[width=\linewidth]{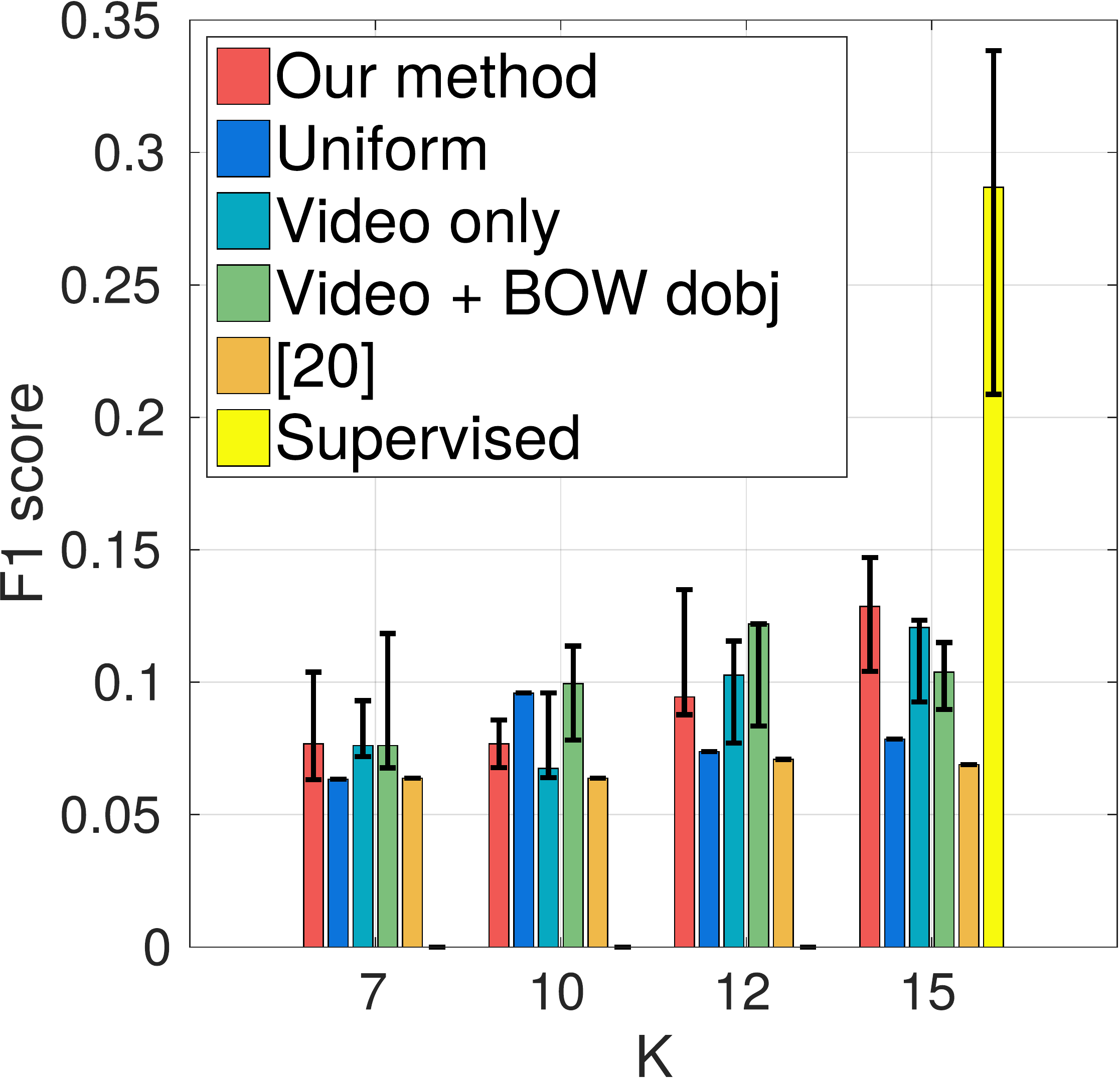}
        \caption{ Jump car (\textbf{12})}
    \end{subfigure}%
    \vspace{-1mm}
    \caption{\small Results for temporally localizing recovered steps in the input videos.
    We give in \textbf{bold} the number of ground truth steps.   
    }
    \vspace{-3mm}
    \label{tab:exp-localization}
\end{figure*}

\section{Experimental evaluation}
\setlength{\tabcolsep}{6pt}
\label{sec:experiments}
In this section, we first describe the details of the text and video features. 
Then we present the results divided into two experiments:
(i) in Sec.~\ref{sec:step_discovery}, we evaluate the quality of steps extracted from video narrations, and
(ii) in Sec.~\ref{sec:localization}, we evaluate the temporal localization of the recovered steps in video using constraints derived from text. 
All the data and code are available at our project webpage~\cite{Alayrac15UnsupervisedWeb}.

\textbf{Video and text features.} 
We represent the transcribed narrations as sequences of direct object relations. 
For this purpose, we run a dependency parser~\cite{Marneffe06generating} on each transcript. 
We lemmatize all direct object relations and keep the ones for which the direct object corresponds to nouns.
To represent a video, we use motion descriptors in order to capture actions (loosening, jacking-up, giving compressions) and frame appearance descriptors to capture the depicted objects (tire, jack, car). 
We split each video into 10-frame time intervals and represent each interval by its motion and appearance descriptors aggregated over a longer block of 30 frames. The motion representation is a histogram of local optical flow (HOF) descriptors aggregated into a single bag-of-visual-word vector of 2,000 dimensions~\cite{Wang13action}.  The visual vocabulary is generated by k-means on a separate large set of training descriptors.
To capture the depicted objects in the video, we apply the VGG-verydeep-16 CNN~\cite{Simonyan14c} over each frame in a sliding window manner over multiple scales. This can be done efficiently in a fully convolutional manner. The resulting 512-dimensional feature maps of conv5 responses are then aggregated into a single bag-of-visual-word vector of 1,000 dimensions, which aims to capture the presence/absence of different objects within each video block. A similar representation (aggregated into compact VLAD descriptor) was shown to work well recently for a variety of recognition tasks~\cite{Cimpoi15}. The bag-of-visual-word vectors representing the motion and the appearance are normalized using the Hellinger normalization and then concatenated into a single 3,000 dimensional vector representing each time interval.

\textbf{WordNet distance.}
For the multiple sequence alignment presented in Sec.~\ref{subsec:model_text}, we set $c(d_1,d_2)=-1$ if $d_1$ and $d_2$ have both their verbs and direct objects that match exactly in the Wordnet tree (distance equal to 0).
Otherwise we set $c(d_1,d_2)$ to be 100.
This is to ensure a high precision for the resulting alignment. %

\subsection{Results of step discovery from text narrations}
\label{sec:step_discovery}
\label{subsec:exp0}

Results of discovering the main steps for each task from text narrations are presented in Table~\ref{exp:MSARes}.
We report results of the multiple sequence alignment described in Sec.~\ref{subsec:model_text} when the maximum number of recoverable steps is $K=10$. Additional results for different choices of $K$ are given in the Appendix~\ref{subsec:script_disc}. %
With increasing $K$, we tend to recover more complete sequences at the cost of occasional repetitions, e.g.~{\em position jack} and {\em jack car} that refer to the same step. 
To quantify the performance, we measure precision as the proportion of correctly recovered steps appearing in the correct order. 
We also measure recall as the proportion of the recovered ground truth steps. 
The values of precision and recall are given at the bottom of Table~\ref{exp:MSARes}.
\subsection{Results of localizing instruction steps in video}
\label{sec:localization}

In the previous section, we have evaluated the quality of the sequences of steps recovered from the transcribed narrations. 
In this section, we evaluate how well we localize the individual instruction steps in the video by running our two-stage approach from Sec.~\ref{subsec:model_var}.

\textbf{Evaluation metric.}
To evaluate the temporal localization, we need to have a one-to-one mapping between the discovered steps in the videos and the ground truth steps.  Following~\cite{Liao05Clustering}, we look for a one-to-one global matching (shared across all videos of a given task) that maximizes the evaluation score for a given method (using the Hungarian algorithm).
Note that this mapping is used only for evaluation, the algorithm does not have access to the ground truth annotations for learning.

The goal is to evaluate whether each ground truth step has been correctly localized in all instruction videos. We thus use the \emph{F1 score} that combines precision and recall into a single score as our evaluation measure. 
For a given video and a given recovered step, our video clustering method predicts exactly one video time interval $t$. 
This detection is considered correct if the time interval falls inside any of the corresponding ground truth intervals, and incorrect otherwise (resulting in a false positive for this video). 
We compute the recall across all steps and videos, defined as the ratio of the number of correct predictions over the total number of possible ground truth steps across videos. A recall of 1 indicates that every ground truth step has been correctly detected across all videos. 
The recall decreases towards 0 when we miss some ground truth steps (missed detections). 
This happens either because this step was not recovered globally, or because it was detected in the video at an incorrect location. This is because the algorithm predicts exactly one occurrence of each step in each video. %
Similarly, precision measures the proportion of correct predictions among all  $N \! \cdot \!K_{\mathrm{pred}}$  possible predictions, where $N$ is the number of videos and $K_{\mathrm{pred}}$
is the number of main steps used by the method. The F1 score is the harmonic mean of precision and recall, giving a score that ranges between 0 and 1, with the perfect score of 1 when all the steps are predicted at their correct locations in all videos.  

\textbf{Hyperparameters.} 
We set the values of parameters $\Delta_b$ and $\Delta_a$ to 0 and 10 seconds. %
The setting is the same for all five tasks.
This models the fact that typically each step is first described verbally and then performed on the camera.
We set $\lambda = 1/(N K_{\mathrm{pred}})$ for all methods that use~\eqref{eq:discriminative}.

\textbf{Baselines.} 
We compare results to four baselines. 
To demonstrate the difficulty of our dataset, we first evaluate a ``Uniform" baseline, which simply distributes instructions steps uniformly over the entire instruction video.
The second baseline ``Video only"~\cite{Bojanowski14weakly} does not use the narration and performs only discriminative clustering on visual features with a global order constraint.\footnote{We use here the improved model from~\cite{Bojanowski15weakly} which does not require a ``background class'' and yields a stronger baseline equivalent to our model~\eqref{eq:videocost} \emph{without} the weak textual constraints.}
The third baseline ``Video + BOW dobj'' basically adds text-based features to the ``Video only'' baseline (by concatenating the text and video features in the discriminative clustering approach).
Here the goal is to evaluate the benefits of our two-stage clustering approach, in contrast to this single-stage clustering baseline. 
The text features are bag-of-words histograms over a fixed vocabulary of direct object relations.\footnote{Alternative features of bag-of-words histograms treating separately nouns and verbs also give similar results.}
The fourth baseline is our own implementation of the alignment method of~\cite{Malmaud15what} (without the supervised vision refinement procedure that requires a set of pre-trained visual classifiers that are not available a-priori in our case).
We use~\cite{Malmaud15what} to re-align the speech transcripts to the sequence of steps discovered by our method of Sec.~\ref{subsec:model_text} (as a proxy for the recipe assumed to be known in~\cite{Malmaud15what}).\footnote{Note that our method finds at the same time the sequence of steps (a recipe in ~\cite{Malmaud15what}) and the alignment of the transcripts.}
To assess the difficulty of the task and dataset, we also compare results with a ``Supervised" approach.
The classifiers $W$ for the visual steps are trained by running the discriminative clustering of Sec.~\ref{sec:model-video} with only ground truth annotations as constraints on the training set.
At test time, these classifiers are used to make predictions under the global ordering constraint on unseen videos.
We report results using 5-fold cross validation for the supervised approach, with the variation 
across folds giving the error bars. 
For the unsupervised discriminative clustering methods, the error bars represent the variation of performance obtained from different rounded solutions collected during the Frank-Wolfe optimization.

\begin{figure}
\centering\includegraphics[width=0.98\linewidth]{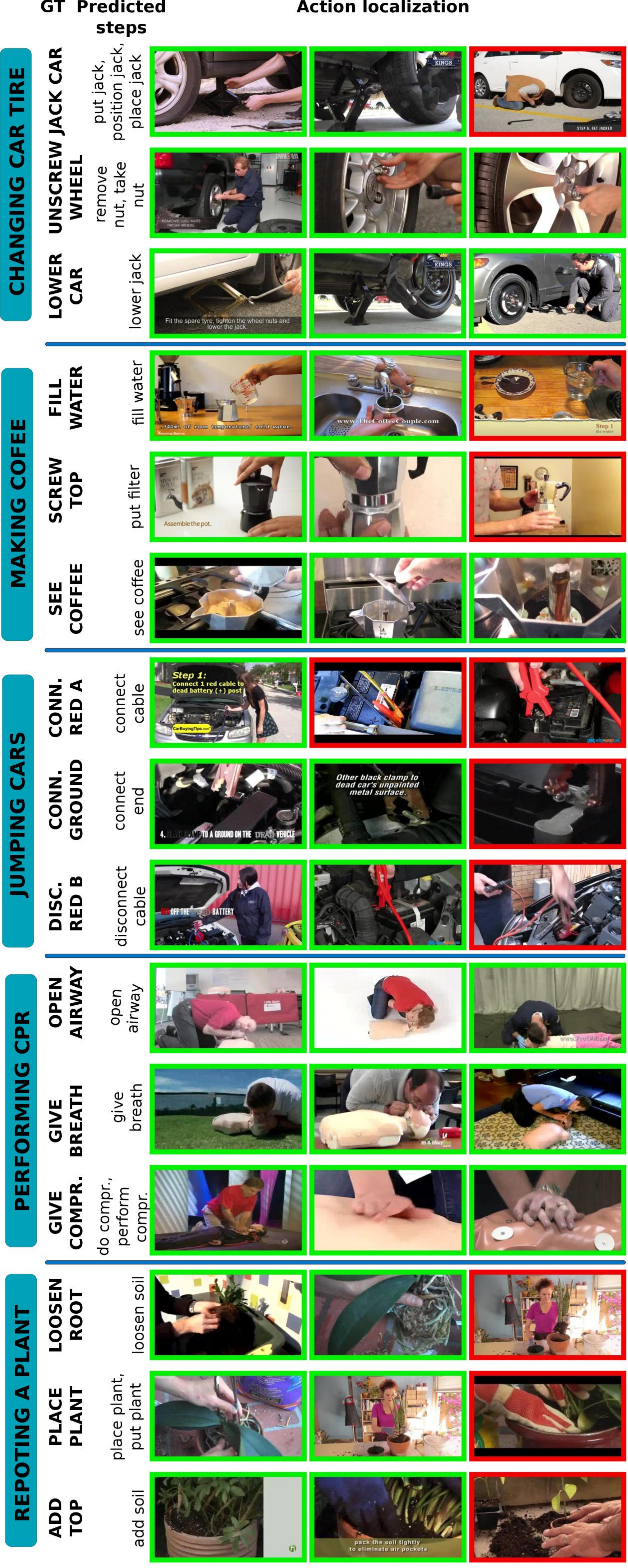}
\vspace{-1mm}
\caption{\footnotesize {\bf Examples of three recovered instruction steps for each of the five tasks in our dataset.} %
For each step, we first show clustered direct object relations, followed by representative example frames localizing the step in the videos. Correct localizations are shown in green. Some steps are incorrectly localized in some videos (red), but often look visually very similar.  
{\bf See Appendix~\ref{subsec:action_loc} for additional results.}
}
\label{fig:qualitative_results} 
\end{figure}

\textbf{Results.} 
Results for localizing the discovered instruction steps are shown
in Figure~\ref{tab:exp-localization}. 
In order to perform a fair comparison to the baseline methods that require a known number of steps $K$, we report results for a range of $K$ values. Note that in our case the actual number of automatically recovered steps can be (and often is) smaller than $K$.
For {\em Change tire} and {\em Perform CPR}, our method consistently outperforms all baselines for all values of $K$ demonstrating the benefits of our approach. 
For {\em Repot}, our method is comparable to text-based baselines, underlying the importance of the text signal for this problem.
For {\em Jump car}, our method delivers the best result (for $K=15$) but struggles for lower values of $K$, which we found was due to visually similar repeating steps (e.g. start car A and start car B) which are mixed-up for lower values of $K$.
For the {\em Make coffee} task, the video only baseline is comparable to our method, which by inspecting the output could be attributed to large variability of narrations for this task. 
Qualitative results of the recovered steps are illustrated in Figure~\ref{fig:qualitative_results}.

\section{Conclusion and future work}
We have described a method to automatically discover the main steps of a task from a set of narrated instruction videos in an unsupervised manner.
The proposed approach has been tested on a new annotated dataset of challenging real-world instruction videos containing complex person-object interactions in a variety of indoor and outdoor scenes. Our work opens up the possibility for large scale learning from instruction videos on the Internet.
Our model currently assumes the existence of a common script with a fixed ordering of the main steps.
While this assumption is often true, e.g.~one cannot remove the wheel before jacking up the car, or make coffee before filling the water, some tasks can be performed while swapping (or even leaving out) some of the steps. Recovering more complex temporal structures is an interesting direction for future work. 

\small{
\paragraph{Acknowledgments}
This research was supported in part by a Google Research Award, and the ERC grants VideoWorld (no. 267907), Activia (no. 307574) and LEAP (no. 336845).
}

{\small
\bibliographystyle{ieee}
\bibliography{biblio}

\begin{thebibliography}{10}\itemsep=-1pt

\bibitem{Alayrac15UnsupervisedWeb}
Project webpage (code/dataset).
\newblock \url{http://www.di.ens.fr/willow/research/instructionvideos/}.

\bibitem{Bach07diffrac}
F.~Bach and Z.~Harchaoui.
\newblock {DIFFRAC}: {A} discriminative and flexible framework for clustering.
\newblock In {\em NIPS}, 2007.

\bibitem{Bojanowski13finding}
P.~Bojanowski, F.~Bach, I.~Laptev, J.~Ponce, C.~Schmid, and J.~Sivic.
\newblock Finding actors and actions in movies.
\newblock In {\em ICCV}, 2013.

\bibitem{Bojanowski14weakly}
P.~Bojanowski, R.~Lajugie, F.~Bach, I.~Laptev, J.~Ponce, C.~Schmid, and
  J.~Sivic.
\newblock Weakly supervised action labeling in videos under ordering
  constraints.
\newblock In {\em ECCV}, 2014.

\bibitem{Bojanowski15weakly}
P.~Bojanowski, R.~Lajugie, E.~Grave, F.~Bach, I.~Laptev, J.~Ponce, and
  C.~Schmid.
\newblock Weakly-supervised alignment of video with text.
\newblock In {\em ICCV}, 2015.

\bibitem{Chambers08}
N.~Chambers and D.~Jurafsky.
\newblock Unsupervised learning of narrative event chains.
\newblock In {\em ACL}, 2008.

\bibitem{Cimpoi15}
M.~Cimpoi, S.~Maji, and A.~Vedaldi.
\newblock Deep filter banks for texture recognition and segmentation.
\newblock In {\em CVPR}, 2015.

\bibitem{Marneffe06generating}
M.-C. de~Marneffe, B.~MacCartney, and C.~D. Manning.
\newblock Generating typed dependency parses from phrase structure parses.
\newblock In {\em LREC}, 2006.

\bibitem{Duchenne2009automatic}
O.~Duchenne, I.~Laptev, J.~Sivic, F.~Bach, and J.~Ponce.
\newblock Automatic annotation of human actions in video.
\newblock In {\em ICCV}, 2009.

\bibitem{Fellbaum98Wordnet}
C.~Fellbaum.
\newblock Wordnet: An electronic lexical database.
\newblock {\em Cambridge, MA: MIT Press.}, 1998.

\bibitem{Frermann14}
L.~Frermann, I.~Titov, and M.~Pinkal.
\newblock A hierarchical \uppercase{B}ayesian model for unsupervised induction
  of script knowledge.
\newblock In {\em EACL}, 2014.

\bibitem{Higgins88clustal}
D.~G.~Higgins and P.~M.~Sharp.
\newblock Clustal: {A} package for performing multiple sequence alignment on a
  microcomputer.
\newblock {\em Gene}, 1988.

\bibitem{Jaggi2013}
M.~Jaggi.
\newblock Revisiting {F}rank-{W}olfe: Projection-free sparse convex
  optimization.
\newblock In {\em ICML}, 2013.

\bibitem{Joulin14efficient}
A.~Joulin, K.~Tang, and L.~Fei-Fei.
\newblock Efficient image and video co-localization with {F}rank-{W}olfe
  algorithm.
\newblock In {\em ECCV}, 2014.

\bibitem{Lacoste15GlobalLinearFW}
S.~Lacoste-Julien and M.~Jaggi.
\newblock On the global linear convergence of {F}rank-{W}olfe optimization
  variants.
\newblock In {\em NIPS}, 2015.

\bibitem{Laptev08a}
I.~Laptev, M.~Marszalek, C.~Schmid, and B.~Rozenfeld.
\newblock Learning realistic human actions from movies.
\newblock In {\em CVPR}, 2008.

\bibitem{Lee01poa}
C.~Lee, C.~Grasso, and M.~Sharlow.
\newblock Multiple sequence alignment using partial order graphs.
\newblock {\em Bioinformatics}, 2002.

\bibitem{Liao05Clustering}
T.~Liao.
\newblock Clustering of time series data, a survey.
\newblock {\em Pattern recognition}, 2014.

\bibitem{Malmaud15what}
J.~Malmaud, J.~Huang, V.~Rathod, N.~Johnston, A.~Rabinovich, and K.~Murphy.
\newblock What's cookin'? {I}nterpreting cooking videos using text, speech and
  vision.
\newblock In {\em NAACL}, 2015.

\bibitem{Miller95Wordnet}
G.~A. Miller.
\newblock Wordnet: A lexical database for english.
\newblock {\em Communications of the ACM}, 1995.

\bibitem{Naim15discriminative}
I.~Naim, Y.~Chol~Song, Q.~Liu, L.~Huang, H.~Kautz, J.~Luo, and D.~Gildea.
\newblock Discriminative unsupervised alignment of natural language
  instructions with corresponding video segments.
\newblock In {\em NAACL}, 2015.

\bibitem{Niebles10a}
J.~C. Niebles, C.-W. Chen, and L.~Fei-Fei.
\newblock Modeling temporal structure of decomposable motion segments for
  activity classification.
\newblock In {\em ECCV}, 2010.

\bibitem{Niebles08}
J.~C. Niebles, H.~Wang, and L.~Fei-Fei.
\newblock Unsupervised learning of human action categories using
  spatial-temporal words.
\newblock {\em IJCV}, 2008.

\bibitem{Potapov14category}
D.~Potapov, M.~Douze, Z.~Harchaoui, and C.~Schmid.
\newblock Category-specific video summarization.
\newblock In {\em ECCV}, 2014.

\bibitem{Raptis13}
M.~Raptis and L.~Sigal.
\newblock Poselet key-framing: {A} model for human activity recognition.
\newblock In {\em CVPR}, 2013.

\bibitem{Regneri10learning}
M.~Regneri, A.~Koller, and M.~Pinkal.
\newblock Learning script knowledge with {W}eb experiments.
\newblock In {\em ACL}, 2010.

\bibitem{Sener15unsupervised}
O.~Sener, A.~Zamir, S.~Savarese, and A.~Saxena.
\newblock Unsupervised semantic parsing of video collections.
\newblock In {\em ICCV}, 2015.

\bibitem{Simonyan14c}
K.~Simonyan and A.~Zisserman.
\newblock Very deep convolutional networks for large-scale image recognition.
\newblock In {\em ICLR}, 2015.

\bibitem{Sun14ranking}
M.~Sun, A.~Farhadi, and S.~Seitz.
\newblock Ranking domain-specific highlights by analyzing edited videos.
\newblock In {\em ECCV}, 2014.

\bibitem{Wang13action}
H.~Wang and C.~Schmid.
\newblock Action recognition with improved trajectories.
\newblock In {\em ICCV}, 2013.

\bibitem{wang1994msaNPhard}
L.~Wang and T.~Jiang.
\newblock On the complexity of multiple sequence alignment.
\newblock {\em Journal of computational biology}, 1(4):337--348, 1994.

\end{thebibliography}


\begin{thebibliography}{1}\itemsep=-1pt

\bibitem{Chari15FWtrack}
V.~Chari, S.~Lacoste-Julien, I.~Laptev, and J.~Sivic.
\newblock On pairwise costs for network flow multi-object tracking.
\newblock In {\em CVPR}, 2015.

\bibitem{lacoste16nonconvexFW}
S.~Lacoste-Julien.
\newblock Convergence rate of {F}rank-{W}olfe for non-convex objectives.
\newblock {\em arXiv preprint}, 2016.

\end{thebibliography}
}

\clearpage

\appendix

\section*{Outline of Supplementary Material}
\label{app:intro}
This supplementary material provides additional details for our method and presents a more complete set of results.
Section~\ref{app:dataset} gives detailed statistics and an illustration of the newly collected dataset of instruction videos.
Section~\ref{app:text_msa} gives details about our new formulation of the multiple sequence alignment problem (Section~\ref{subsec:model_text} of the main paper) as a quadratic program and presents empirical results showing that our Frank-Wolfe optimization approach obtains solutions with lower objective values than the state-of-the-art heuristic algorithms for multiple sequence alignment. 
Section~\ref{sec:details_diffrac} provides the details for the discriminative clustering of videos with text constraints that was briefly described in Section~\ref{sec:model-video} of the main paper.
Section~\ref{subsec:details_experiments} gives additional details about the experimental protocol used in Section~\ref{sec:localization} in the main paper.
Finally, in Section~\ref{sec:qual_res}, we give a more complete set of qualitative results for both the clustering of transcribed verbal instructions (see~\ref{subsec:script_disc}) and localizing instruction steps in video (see~\ref{subsec:action_loc}).

\section{New challenging dataset of instruction videos}
\label{app:dataset}

\subsection{Dataset statistics}
\label{subsec:score_dataset}

In this section, we introduce three different scores which aim to illustrate different properties of our dataset.
The scores characterize (i) the step ordering consistency, (ii) the missing steps and (iii) the possible step repetitions.
 
Let $N$ be the number of videos for a given task and $K$ the number of steps defined in the ground truth.
We assume that the ground truth steps are given in an ordered fashion, meaning the global order is defined as the sequence $\left\lbrace 1,\ldots, K \right\rbrace $. 
For the $n$-th video, we denote by $g_n$ the total number of annotated steps, by $u_n$ the number of unique annotated steps and finally by $l_n$ the length of the longest common subsequence between the annotated sequence of steps and the ground truth sequence $\left\lbrace 1,\ldots, K \right\rbrace$.

\paragraph{Order consistency error.}
The \emph{order error} score $O$ is defined as the proportion of non repeated annotated steps that are not consistent with the global ordering.
In other words, it is defined as the number of steps that do not fit the global ordering defined in the ground truth divided by the total number of unique annotated steps.
More formally, $O$ is defined as follows:
\begin{equation}
O := 1-\frac{\sum_{n=1}^N l_n}{\sum_{n=1}^N u_n}.
\label{order_score}
\end{equation}

\paragraph{Missing steps.} 
We define the \emph{missing steps} score $M$ as the proportion of steps that are visually missing in the videos when compared to the ground truth.
Formally, 
\begin{equation}
M := 1-\frac{\sum_{n=1}^N u_n}{KN}.
\label{missing_score}
\end{equation}

\paragraph{Repeated steps.}
The \emph{repetition score} $R$ is defined as the proportion of steps that are repeated:
\begin{equation}
R := 1-\frac{\sum_{n=1}^N u_n}{\sum_{n=1}^N g_n}.
\label{repet_score}
\end{equation}

\paragraph{Results.}
In Table~\ref{tab:dataset_stat}, we give the previously defined statistics for the five tasks of the instruction videos dataset.
Interestingly, we observed that globally the order is consistent for the five tasks with a total order error of only $6\%$.
Steps are missing in $27\%$ of the cases. 
This illustrates the difficulty of defining the right granularity of the ground truth for this task. 
Indeed, some steps might be optional and thus not visually demonstrated in all videos.
Finally the global repetition score is $14\%$.
Looking more closely, we observe that the \emph{Performing CPR} task is the main contributor to this score.
This is obviously a good example where one needs to repeat several times the same steps (here alternating between compressions and giving breath).
Even if our model is not explicitly handling this case, we observed that our multiple sequence alignment technique for clustering the text inputs discovered these repetitions (see Table~\ref{tab:script_res}).
Finally, these statistics show that the problem introduced in this paper is very challenging and that designing models which are able to capture more complex structure in the organization of the steps is a promising direction for future work.

\begin{table*}[ht]
\centering
\begin{tabular}{@{}lrrrrrr@{}}
\toprule
Task             & \multicolumn{1}{l}{Changing tire} & \multicolumn{1}{l}{Performing CPR} & \multicolumn{1}{l}{Repoting plant} & Making coffee & \multicolumn{1}{c}{Jumping cars} & \multicolumn{1}{c}{\textbf{Average}} \\ \midrule
Order error      & 0.7\%                             & 11\%                               & 6\%                                & 3\%           & 8\%                              & \textbf{6\%}                       \\
Missing steps    & 16\%                              & 32\%                               & 30\%                               & 28\%          & 27\%                             & \textbf{27\%}                      \\
Repetition score & 4\%                               & 50\%                               & 7\%                                & 11\%          & 0.4\%                            & \textbf{14\%}                      \\ \bottomrule
\end{tabular}
\caption{Statistics of the instruction video dataset.}
\label{tab:dataset_stat}
\end{table*}

\subsection{Complete illustration of the dataset}

\begin{figure*}[ht!]
       \centering
	 \includegraphics[width=\linewidth]{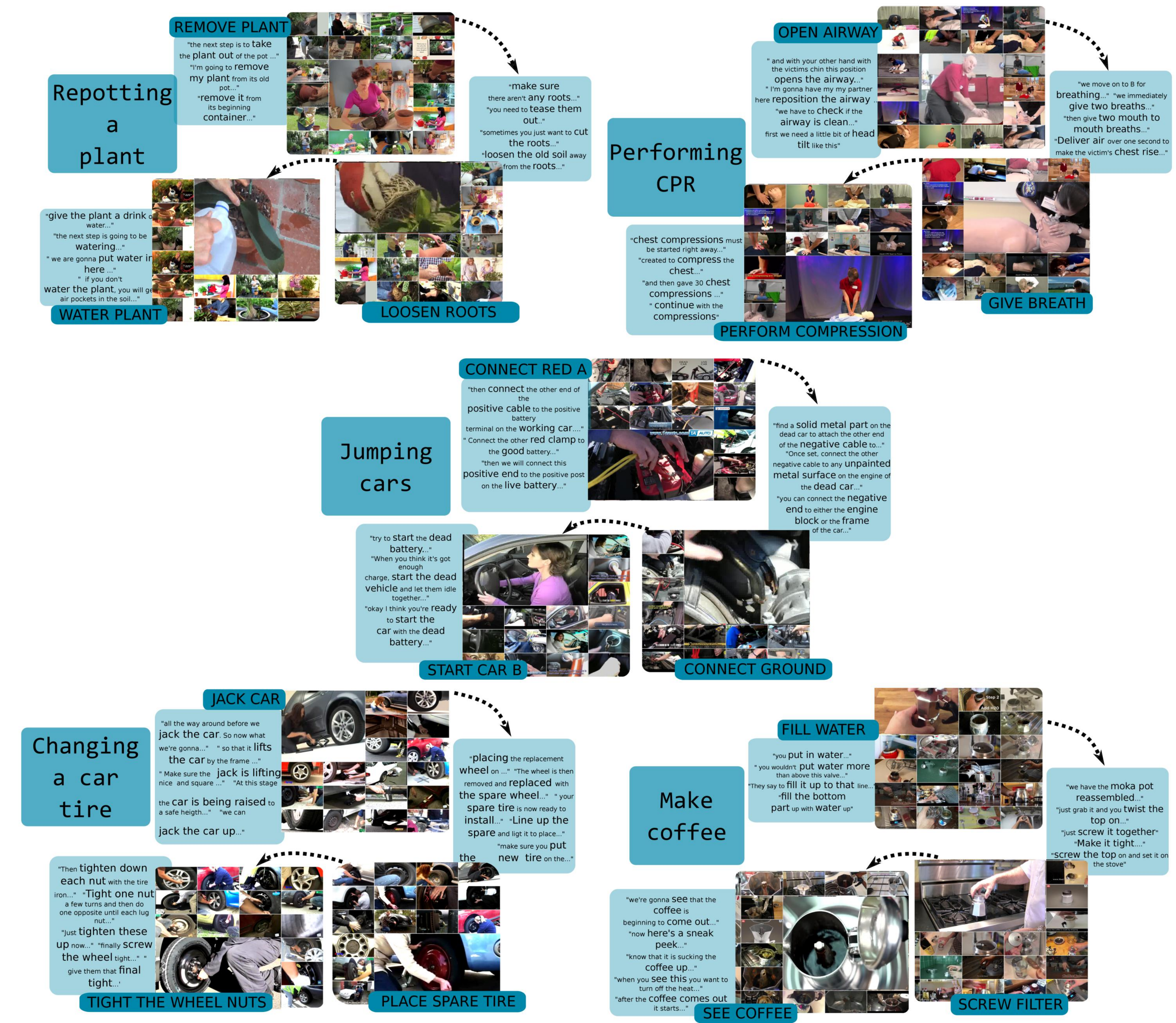} %
     \caption{\small Illustration of our newly collected dataset of instructions videos. 
     Examples of transcribed narrations together with still frames from the corresponding videos are shown for the 5 tasks of the dataset: 
     {\em Repotting a plant}, {\em Performing CPR}, {\em Jumping cars}, {\em Changing a car tire} and {\em Making coffee}. 
     The dataset contains challenging real-world videos performed by many different people, captured in uncontrolled settings in a variety of outdoor and indoor environments.  
     }
     \label{fig:datasetApp}
 \end{figure*}

Figure~\ref{fig:datasetApp} illustrates all five tasks in our newly collected dataset.
For each task, we show a subset of 3 events that compose the task.
Each event is represented by several sample frames and extracted verbal narrations.
Note the large variability of verbal expressions and the terminology in the transcribed narrations as well as the large variability of visual appearance  due to viewpoint, used objects, and actions performed in different manner.
At the same time, note the the consistency of the actions between the different videos and the underlying script of each task.
\section{Clustering transcribed verbal instructions}
\label{app:text_msa}
In this section, we review in details the way we model the text clustering.
In particular, we give details on how we can reformulate multiple sequence alignment as a quadratic program.
Recall that we are given $N$ narrated instruction videos.
For the $n$-th video, the text signal is represented as a sequence of direct object relation tokens : $d^n=(d_1^n,\dots,d^n_{S_n})$, where the length $S_n$ of the sequences varies from one video clip to another.
The number of possible direct object relations in our dictionary is denoted $D$.
The multiple sequence alignment (MSA) problem was formulated
as mapping each input sequence $d^n$ of tokens to a global common
template of $L$ slots, while minimizing the sum-of-pairs score given in~\eqref{eq:msa_cost}.
For each input sequence $d^n$, we used the notation $(\phi(d^n))_{1\leq l\leq L}$
to denote the re-mapped sequence of tokens into $L$ slots: $\phi(d^n)_l$
represents the direct object relation put at location $l$, with $\phi(d^n)_l = \varnothing$
denoting that a gap was inserted in the original sequence and
the slot $l$ is left empty.
We also have defined a cost $c(d_1, d_2)$ of aligning two direct object 
relations together, with the possibility that $d_1$ or $d_2$
is~$\varnothing$, in which case we defined the cost to be $0$
by default.
In the following, we summarize the cost of aligning non-empty
direct object relations
by the matrix $\Ctext\in\mathbb{R}^{D\times D}$.
$(\Ctext)_{ij}$ is equal to the cost of aligning the $i$-th and the 
$j$-th direct object relation from the dictionary together.

\subsection{Reformulating multiple sequence alignment as a quadratic program}
\label{subsec:details_msa}

We now present our formalization of the search problem as a quadratic program.
To the best of our knowledge this is a new formulation of the multiple sequence alignment (MSA) problem, 
which in our setting (results shown later) consistently obtains better values of the multiple sequence alignment objective than the current
state-of-the-art MSA heuristic algorithms.
 
We encode the identity of a direct object relation with a $D$-dimensional
indicator vector.
The text sequence $n$ can then be represented by an indicator matrix $Y_n \in \{0,1\}^{S_n\times D}$.
The $j$-th row of $Y_n$ indicates which direct object relations is evoked at the $j$-th position.
Similarly, the token re-mapping  $(\phi(d^n))_{1\leq l\leq L}$ can
be represented as a $L \times D$ indicator matrix; where 
each row $l$ encodes which token is appearing in slot $l$ (and
a whole row of zero is used to indicates an empty $\varnothing$ slot).
This re-mapping can be constructed from two 
pieces of information: first, which token index $s$ of the original sequence
is re-mapped to which global template slot $l$; 
we represent this by the decision matrix  $U_n\in\{0,1\}^{S_n\times L}$,
which satisfies very specific constraints (see below).
The second piece of information is the composition 
of the input sequence encoded by $Y_n$.
We thus have $\phi(d^n) = U_n^T Y_n$ (as a $L \times D$ indicator
matrix). Given this encoding, the cost matrix $\Ctext$, and the fact
that the alignment of empty slots has zero cost, we
can then rewrite the MSA problem that minimizes
the sum-of-pairs objective~\eqref{eq:msa_cost}
as follows:
\begin{equation}
\begin{aligned}
& \underset{U_n, n\in\{1,\ldots,N\}}{\text{minimize}}
& & \sum_{(n,m)} \mathrm{Tr}(U_n^TY_n \Ctext Y_m^TU_m) \\
& \text{subject to}
& & U_n \in \mathcal{U}_n, \; n = 1, \ldots, N.
\end{aligned}
\label{eq:msaeq}
\end{equation}
In the above equation, the trace ($\mathrm{Tr}$) is computing the cost of aligning sequence $m$ with sequence $n$
(the inner sum in~\eqref{eq:msa_cost}).
Moreover, $\mathcal{U}_n$ is a constraint set that encodes the fact that $U_n$ has to be a valid (increasing) re-mapping.\footnote{More formally  $\mathcal{U}_n :=\{U\in\{0,1\}^{S_n \times L}$ s.t.  $U\textbf{1}_L=\textbf{1}_{S_n}$ and $\forall l, \left( U_{sl}=1 \Rightarrow ((\forall s'>s,l'\leq l),  U_{s'l'}=0 \right)\}$.}
As before, we can eliminate the video index $n$ by simply stacking the assignment matrices $U_n$ in one matrix $U$ of size $S\times L$.
Similarly, we denote $Y$ the $S\times D$ matrix which is obtained by the concatenation of all the $Y_n$ matrices. 
We can then rewrite the equation~(\ref{eq:msaeq}) as a quadratic program over the (integer) variable $U$:

\begin{equation}
\underset{U}{\text{minimize}} \ \mathrm{Tr}(U^TBU), \ \text{subject to} \ U \in \mathcal{U}.
\label{eq:msaeqcompact}
\end{equation}
In this equation, the $S \times S$ matrix $B$ is deduced from the input sequences and the cost between different direct object relations by computing $B := Y \Ctext Y^T$.
It represents the pairwise cost at the token level, i.e. the cost of aligning token $s$ in one
sequence to token $s'$ in another sequence.

\subsection{Comparison of methods}
\label{subsec:comparison_msa}

\begin{table*}[!ht] 
    \centering
    \setlength{\tabcolsep}{.6em} 
    \begin{tabular}{lccccc} 
        \toprule
        Task 			&  Changing tire   &  Performing CPR  & Repotting plant &  Making coffee   & Jumping cars\\ 
        \midrule
        Poa~\cite{Lee01poa} 			    &   11.30   &   -3.82   &  1.65    &   -2.99   &   4.55	\\ 
        Ours using Frank-Wolfe 					    &  \textbf{-5.18}  &   \textbf{-4.51}   & \textbf{-3.55} &   \textbf{-3.86}   & \textbf{-4.67} 	\\ 
        \bottomrule
    \end{tabular}  
    \caption{Comparison of different optimization approaches for solving problem~\eqref{eq:msaeqcompact}. (Objective value, lower is better).}  
    \label{tab:msaObj}
\end{table*} 

The problem~(\ref{eq:msaeqcompact}) is NP-hard~\cite{wang1994msaNPhard} in general, as is typical 
for integer quadratic programs.
However, much work has been done in computational biology
to develop efficient heuristics to solve the MSA problem,
as it is an important problem in their field.
We briefly describe below some of the existing heuristics to solve it,
and then present our Frank-Wolfe optimization approach,
which gave surprisingly good empirical results
for our problem.\footnote{We stress here that we do not claim 
that our formulation of the multiple sequence alignment (MSA) problem as a quadratic program outperforms the state-of-the-art 
computational biology heuristics for their MSA problems \emph{arising in biology}.
We report our observations on application of multiple sequence alignment to our application,
which might have a structure for which these heuristics are not as appropriate.}

\paragraph{Standard methods.}
Here, we compare to a standard state-of-the-art method for multiple sequence alignment~\cite{Lee01poa}.
Similarly to~\cite{Higgins88clustal}, they first align two sequences and merge them in a common template.
Then they align a new sequence to the template and then update the template.
They continue like this until no sequence is left.
Differently from~\cite{Higgins88clustal}, they use a better representation of the template by using partial order graph instead of simple linear representations.
This gives more accuracy for the final alignment.
For the experiments, we use the author's implementation.\footnote{Code available at \url{http://sourceforge.net/projects/poamsa/}.}

\paragraph{Our solution using Frank-Wolfe optimization.}
We first note that problem~\eqref{eq:msaeqcompact} has a very similar 
structure to an optimization problem that we solve using Frank-Wolfe optimization 
for the discriminative clustering of videos; see Equations~\eqref{eq:appAexph} and~\eqref{eq:hFWproblem} below.
For this, we first perform a continuous relaxation of the set of constraints $\mathcal{U}$ by replacing it with its convex hull $\bar{\mathcal{U}}$.
The Frank-Wolfe optimization algorithm~\cite{Jaggi2013} can solve quadratic program over constraint sets
for which we have access to an efficient linear minimization oracle.
In the case of $\mathcal{U}$,
the linear oracle can be solved exactly with a dynamic program very similar to the one described in Section~\ref{subsec:fw_dp}.
We note here that even with the continuous relaxation over $\bar{\mathcal{U}}$, the resulting problem is still non-convex because $B$ is not positive semidefinite -- this is because of the cost function appearing in the MSA problem.
However, the standard convergence proof for Frank-Wolfe can easily be extended to show
that it converges at a rate of $O(1/\sqrt{k})$ to a stationary point on non-convex objectives~\citesup{lacoste16nonconvexFW}.
Once the algorithm has converged to a (local) stationary point,
we need to round the fractional solution to obtain a valid encoding $U$.
We follow here a similar rounding strategy that was originally proposed by~\citesup{Chari15FWtrack}
and then re-used in~\cite{Joulin14efficient}: we pick the last visited
corner (which is necessarily integer) 
which was given as a solution to the linear minimization oracle
(this is called Frank-Wolfe rounding). 

\paragraph{Results.} In Table~\ref{tab:msaObj}, we give the value of the objective~(\ref{eq:msaeqcompact}) for the rounded solutions obtained by the two different optimization approaches (lower is better), for the MSA problem on our
five tasks.
Interestingly, we observe that the Frank-Wolfe algorithm consistently outperforms the state-of-the-art method of~\cite{Lee01poa} in our setting.

\section{Discriminative clustering of videos under text constraints }
\label{sec:details_diffrac}

We give more details here on the discriminative clustering
framework from~\cite{Bojanowski14weakly,Bojanowski15weakly}
(and our modifications to include the text constraints) that we use
to localize the main actions in the video signal.

\subsection{Explicit form of $h(Z)$}
\label{subsec:explicit_h}

We recall that $h(Z)$ is the cost of clustering all the video streams $\{x^n\}, n=1,\ldots, N$, into a sequence of $K$ steps. 
The design matrix $X$  $\in \mathbb{R}^{T \times d}$ contains the feature describing the time intervals in our videos.
The indicator latent variable $Z\in \mathcal{Z} := \{0,1\}^{T\times K}$ encodes the visual presence of a step $k$ at a time interval $t$. 
Recall also that $X$ and $Z$ contains the information about all videos $n \in \{1,\ldots, N\}$.
Finally, $W\in \mathbb{R}^{d\times K}$ represents a linear classifier for our $K$ steps, that is shared among all videos. We now derive the explicit form of $h(Z)$ as in the DIFFRAC approach~\cite{Bach07diffrac}, though yielding a somewhat simpler expression (as in~\cite{Bojanowski15weakly}) due to our use of a (weakly regularized) bias feature in $X$ instead of a separate (unregularized) bias $b$. 
Consider the following joint cost function $f$ on $Z$ and $W$ defined as
\begin{align}
f(Z,W) =  \frac{1}{2T} \|Z - X W\|_F^2+ \frac{\lambda}{2} \|W\|_F^2.
\label{eq:appAf}
\end{align}
The cost function $f$ simply represents the ridge regression objective with output labels $Z$ and input design matrix $X$. We note that $f$ has the nice property of being \emph{jointly} convex in both $Z$ and $W$, implying that its unrestricted minimization with respect to $W$ yields a \emph{convex} function in $Z$. This minimization defines our clustering cost $h(Z)$; rewriting the definition of $h$ with the joint cost $f$ from~\eqref{eq:appAf}, we have:
\begin{align}
    h(Z) = \min_{W \in \mathbb{R}^{d \times K}} \ f(Z,W).
\label{eq:appAh}
\end{align}
As $f$ is strongly convex in $W$ (for any $Z$), we can obtain its unique minimizer $W^*(Z)$ as a function of $Z$ by zeroing its gradient and solving for $W$.
For the case of the square loss in equation~\eqref{eq:appAf}, the optimal classifier $W^*(Z)$ can be computed in closed form:
\begin{align}
W^*(Z) =  (X^TX+T\lambda I_d)^{-1}X^TZ, 
\label{eq:appAWopt}
\end{align}
where $I_d$ is the $d$-dimensional identity matrix.
We obtain the explicit form for $h(Z)$ by substituting the expression~\eqref{eq:appAWopt} for $W^*(Z)$ in equation~\eqref{eq:appAf} and properly simplifying the expression:
\begin{align}
h(Z) = f(Z,W^*) = \frac{1}{2T}\text{Tr}(ZZ^TB),
\label{eq:appAexph}
\end{align}
where $B := I_T-X(X^TX+T\lambda I_d)^{-1}X^T$ is a strictly positive definite matrix (and so $h$ is actually strongly convex). The clustering cost is a quadratic function in $Z$, encoding how the clustering decisions in one interval $t$ interact with the clustering decisions in another interval $t'$. In the next section, we explain how we can optimize the clustering cost $h(Z)$ subject to the constraints from Section~\ref{sec:model-video} using the Frank-Wolfe algorithm.

\subsection{Frank Wolfe algorithm for minimizing $h(Z)$}
\label{subsec:fw_dp}

The localization of steps in the video stream is done by solving the following optimization problem
(repeated from~\eqref{eq:videocost} here for convenience):
\begin{equation}
\underset{Z}{\text{minimize}} \quad h(Z) \quad  \text{ s.t. }  \underbrace{Z \in \mathcal{Z}}_{\text{ordered script}}, \quad
\underbrace{AZ \geq R}_{\substack{\text{weak textual}\\ \text{constraints}} }.
\label{eq:hFWproblem}
\end{equation}
where $Z$ is the latent assignment matrix of video time intervals to $K$ clusters and $R$ is the matrix of assignments of direct object relations in text to $K$ clusters.
Note that $R$ is obtained from the text clustering using multiple sequence alignment as described in Section~\ref{sec:text} and~\ref{subsec:details_msa}, and is fixed before optimizing over $Z$.
$R$ is a $S \times K$ matrix obtaining by picking the $K$ main columns of the $U$ matrix
defined in Section~\ref{subsec:details_msa}. This selection step
was described in the ``extracting the main steps'' paragraph in Section~\ref{sec:text}.

The constraint set encodes several concepts.
First, it imposes the temporal consistency between the text stream and the video stream. 
We recall that this constraint was written as $AZ \geq R$,\footnote{
When $R_{sk} = 0$, then this constraint does not do anything. When $R_{sk} = 1$ (i.e. the
text token $s$ was assigned to the main action $k$), then the constraint enforces
that $\sum_{t \in A_{s \cdot}} Z_{tk} \geq 1$, where $A_{s \cdot}$ represents
which video frames are temporally close to the caption time of the text token~$s$.
It thus then enforces that at least one temporally close video frame is assigned
to the main action $k$.
} 
where $A$ encodes the temporal alignment constraints between video and text (type I).
Second, it includes the event ordering constraints within each video input (type II).
Finally, it encodes the fact that each event is assigned to exactly one time interval within each video (type III).
The last two constraints are encoded in the set of constraints $\mathcal{Z}$.
To summarize, let $\mathcal{\tilde{Z}}$ denote the resulting (discrete) feasible space for $Z$ i.e. $\mathcal{\tilde{Z}} := \{Z \in \mathcal{Z} \, | \,AZ \geq R \}$.
We are then left with a problem in $Z$ which is still hard to solve because the set $\tilde{\mathcal{Z}}$ is not convex. 
To approximately optimize $h$ over $\tilde{\mathcal{Z}}$, we follow the strategy of~\cite{Bojanowski14weakly,Bojanowski15weakly}.  First, we optimize $h$ over the relaxed $\text{conv}(\tilde{\mathcal{Z}})$ by using the Frank-Wolfe algorithm to get a fractional solution $Z^* \in \text{conv}(\tilde{\mathcal{Z}})$. We then find a feasible candidate $\hat{Z} \in \tilde{\mathcal{Z}}$ by using a rounding procedure. We now give the details of these steps.  

First we note that the linear oracle of the Frank-Wolfe algorithm can be solved separately for each video $n$.
Indeed, because we solve a linear program, there is no quadratic term that brings dependence between different videos in the objective, and moreover all the constraints are blockwise in $n$.
Thus, in the following, we will give details for one video only by adding an index $n$ to $\tilde{\mathcal{Z}}$, to $Z$ and to $T$.

\begin{figure}[tbp]
    \centering
    \includegraphics[width=\linewidth]{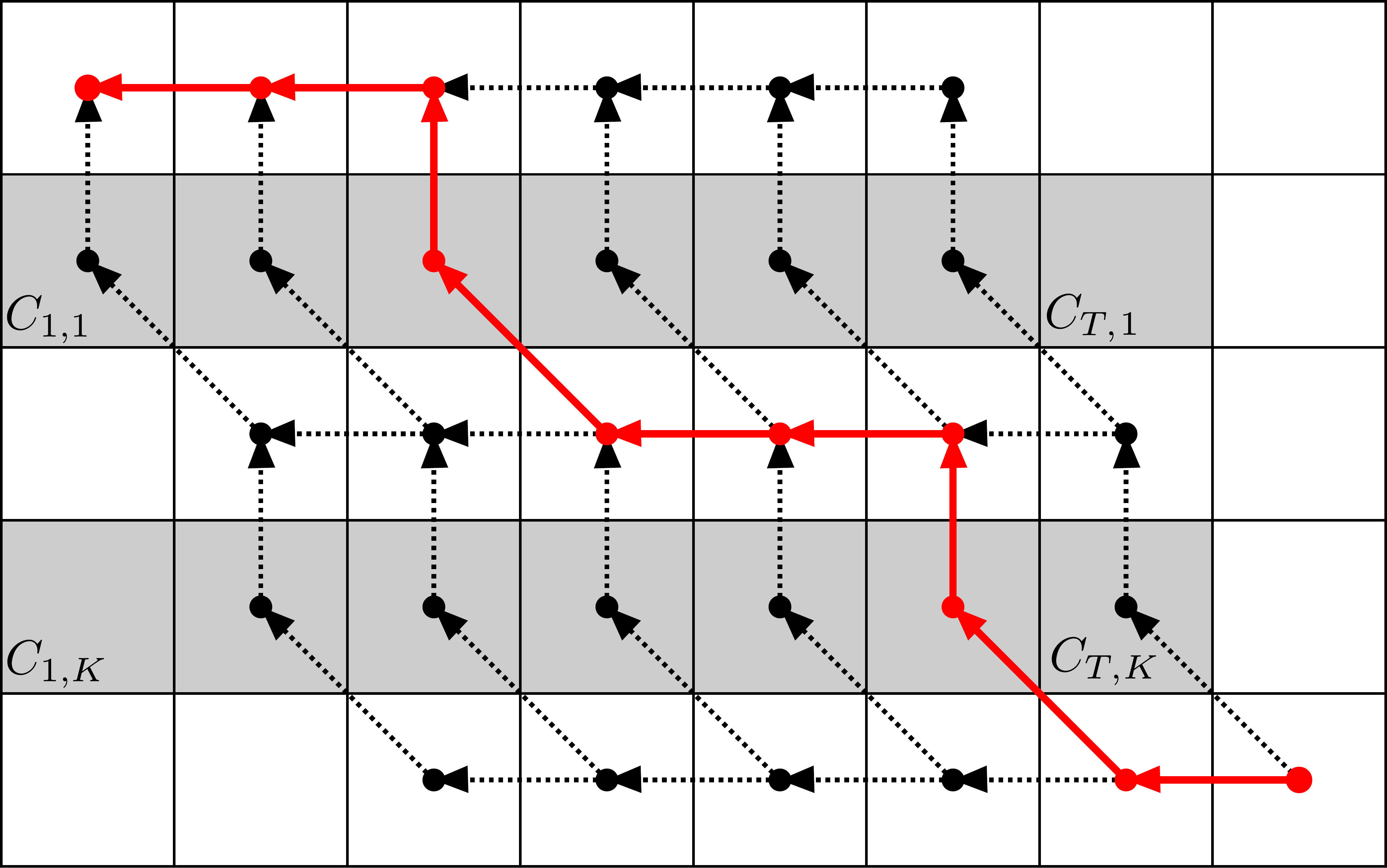}

    \caption{
        Illustration of the dynamic programming solution to the linear program~\eqref{eq:linear}.
        The drawing shows a possible cost matrix~$\tilde{C}$ and an optimal path in red.
        The gray entries in the matrix $\tilde{C}$ correspond to the values from the matrix \(C\). The white entries have minimal cost and are thus always preferred over any gray entry. 
        Note that we display~$\tilde{C}$ in a transpose manner to better fit on the page.
}
    \label{fig:dp}

\end{figure}

The linear oracle of the Frank-Wolfe algorithm can be solved via an efficient dynamic program.
Let us suppose that the linear oracle corresponds to the following problem: 
\begin{equation}
    \min_{Z_n \in \tilde{\mathcal{Z}}_n} \text{Tr}(C_n^\top Z_n),
    \label{eq:linear}
\end{equation}
where $C_n \in \mathbb{R}^{T_n \times K}$ is a cost matrix that arises by computing the gradient of $h$ with respect to $Z_n$ at the current iterate. The goal of the dynamic program is to find which entries of $Z_n$ are equal to 1, recalling that $(Z_n)_{tk}=1$ means that the step $k$ was assigned to time interval $t$. From the constraint of type III (unique prediction per step), we know that each column $k$ of $Z_n$ has exactly one $1$ (to be found). From the ordering constraint (type II), we know that if $(Z_n)_{tk} = 1$, then the only possible locations for a 1 in the $(k+1)$-th column is for $t' > t$ (i.e. the pattern of 1's is going downward when traveling from left to right in $Z_n$). Note that there can be ``jumps'' in between the time assignment for two subsequent steps $k$ and $k+1$. In order to encode this possibility using a continuous path search in a matrix, we insert dummy columns into the cost matrix $C$.
We first subtract the minimum value from $C$ and then insert columns filled with zeros in between every pair of columns of $C$.
In the end, we pad $C$ with an additional row filled with zeros at the bottom.
The resulting cost matrix $\tilde{C}$ is of size $(T_n+1) \times (2K+1)$ and is illustrated (as its transpose) along with the corresponding update rules in Figure~\ref{fig:dp}.

The problem that we are interested in is subject to the additional linear constraints given by the clustering of text transcripts (constraints of type I).
These constraint can be added by constraining the path in the dynamic programming algorithm.
This can be done for instance by setting an infinite alignment cost outside of the constrained region.

At the end of the Frank-Wolfe optimization algorithm, we obtain a continuous solution $Z^*_n$ for each $n$.
By stacking them all together again, we obtain a continuous solution $Z^*$. From the definition of $h$, we can also look at the corresponding model $W^*(Z^*)$ defined by equation~\eqref{eq:appAWopt} which again is shared among all videos.
All $Z_n^*$ have to be rounded in order to obtain a feasible point for the initial, non relaxed problem.
Several rounding options were suggested in~\cite{Bojanowski15weakly}; it turns out that the one which uses $W^*$ gives better results in our case.
More precisely, in order to get a good feasible binary matrix $\hat{Z}_n \in \tilde{Z}_n$, we solve the following problem: $\min_{Z_n \in \tilde{\mathcal{Z}}_n} \ \|Z_n - X_n W^*\|_F^2$.
By expanding the norm, we notice that this corresponds to a simple linear program over $\tilde{\mathcal{Z}}_n$ as in equation~\eqref{eq:linear} that can be solved using again the same dynamic program detailed above.
Finally, we stack these rounded matrices $\hat{Z}_n$ to obtain our predicted assignment matrix $\hat{Z} \in \tilde{\mathcal{Z}}$.

\section{Experimental protocol}
\label{subsec:details_experiments}

In this section, we give more details about the setting for our experiments on the time localization of events with results given in Figure~\ref{tab:exp-localization}.

\subsection{Supervised experiments.} 
Here, we describe in more details how we obtained the scores for the supervised approach depicted in yellow in Figure~\ref{tab:exp-localization}.
We first divided the $N$ input videos in 5 different folds.
One fold is kept for the test set while the 4 other are used as train/validation dataset.
With the 4 remaining folds, we perform a 4-fold cross validation in order to choose the hyperparameter $\lambda$.
Once the hyper parameter is fixed, we retrain a model on the 4 folds and evaluate it on the test set.
By iterating over the five possible test folds, we report variation in performance with error bars in Figure~\ref{tab:exp-localization}.

\paragraph{Training phase.}
The goal of this phase is to learn classifiers $W$ for the visual steps.
To that end, we minimize the cost defined in~\eqref{eq:videocost} under the ground truth annotations constraints.
This is very close to our setting, and in practice we can use exactly the same framework as in problem~\eqref{eq:hFWproblem} by simply replacing the constraints coming from the text by the constraints coming from the ground truth annotations.
\paragraph{Testing phase.}
At test time, we simply use the classifiers $W$ to perform least-square prediction of $Z_{\text{test}}$ under ordering constraints.
Performance are evaluated with the F1 score.

\subsection{Error bars for Frank-Wolfe methods.}
We explain here how we obtained the error bars of Figure~\ref{tab:exp-localization} in the main paper for the unsupervised approaches.
Let us first recall that the Frank-Wolfe algorithm is used to solve a continuous relaxation of problem~\eqref{eq:hFWproblem}.
To obtain back an integer solution, we round the continuous solution using the rounding method described at the end of Section~\ref{subsec:fw_dp}.
This rounding procedure is performed at each iteration of the optimization method.
When the stopping criterion of the Frank-Wolfe scheme is reached (fixed number of iterations or target sub-optimality in practice), we have as many rounded solutions as number of iterations.
Our output integer solution is then the integer point that achieves the lowest objective.
Note that we are only guaranteed to diminish objective in the continuous domain and \emph{not} for the integer points, therefore there are no guarantees that this solution is the last rounded point.
In order to illustrate the variation of the performance with respect to the optimization scheme, we defined our error bars as being the interval with bounds determined by the minimal performance and the maximal performance obtained \emph{after} visiting the best rounded point (the output solution).
This notably explains why the error bars of Figure~\ref{tab:exp-localization} are not necessarily symmetric.
Overall, the observed variation is not very important, thus highlighting the stability of the procedure.

\section{Qualitative results}
\label{sec:qual_res}

In Section~\ref{subsec:script_disc}, we give detailed results of script discovery for the five different tasks.
In Section~\ref{subsec:action_loc}, we present detailed results for the action localization experiment.

\subsection{Script discovery}
\label{subsec:script_disc}
Table~\ref{tab:script_res} shows the automatically recovered sequences of steps for the five tasks considered in this work. 
        The results are shown for setting the maximum number of discovered steps, $K = \{7,10,12,15\}$. 
        Note how our method automatically selects less than $K$ steps in some cases.
         These are the automatically chosen $k\leq K$ steps that are the most salient in the aligned narrations as described in Section~\ref{subsec:model_text}.  
		 This is notably the case for the {\em Repotting a plant} task. 
		 Even for $K\leq 12$, the algorithm recovers only 6 steps that match very well the seven ground truth steps for this task. 
		 This saliency based task selection is important because it allows for a better precision at high $K$ without lowering much the recall.       

Please note also how the steps and their ordering recovered by our method correspond well to the ground truth steps for each task. 
        For {\em CPR}, our method recovers fine-grained steps e.g.~{\em tilt head}, {\em lift chin}, which are not included in the main ground truth steps, but nevertheless could be helpful in some situations.
	    For {\em Changing tire}, we also recover more detailed actions such as~{\em remove jack} or {\em put jack}.
	    In some cases, our method recovers repeated steps. %
	    For example, for {\em CPR} our method learns that one has to alternate between {\em giving breath} and {\em performing compressions} even if this alternation was not annotated in the the ground truth.
	     Or for  {\em Jumping Cars} our method learns that cables need to be connected twice (to both cars).

	    These results demonstrate that our method is able to automatically discover meaningful scripts describing very different tasks.
	    The results also show that the constraint of a single script providing an ordering of events is a reasonable prior for a variety of different tasks. 

\subsection{Action localization}
\label{subsec:action_loc}
\begin{figure*}[!ht]
\includegraphics[width=\linewidth]{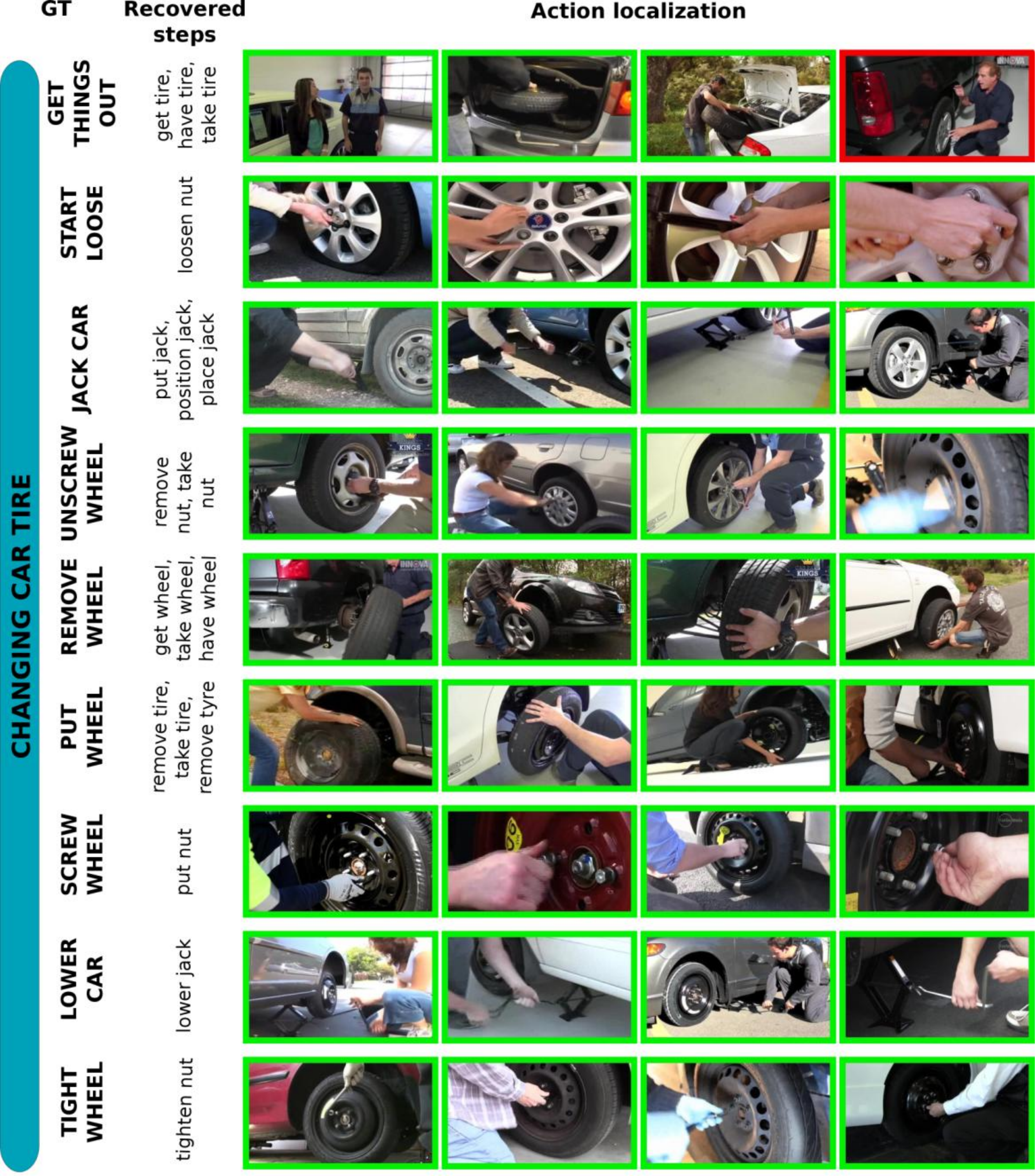}
\caption{\footnotesize {\bf Examples of the recovered instruction steps for the task ``Changing the car tire".} 
}

\label{fig:qualitative_results_changing_tire} 
\end{figure*}

\begin{figure*}[!ht]
\includegraphics[width=\linewidth]{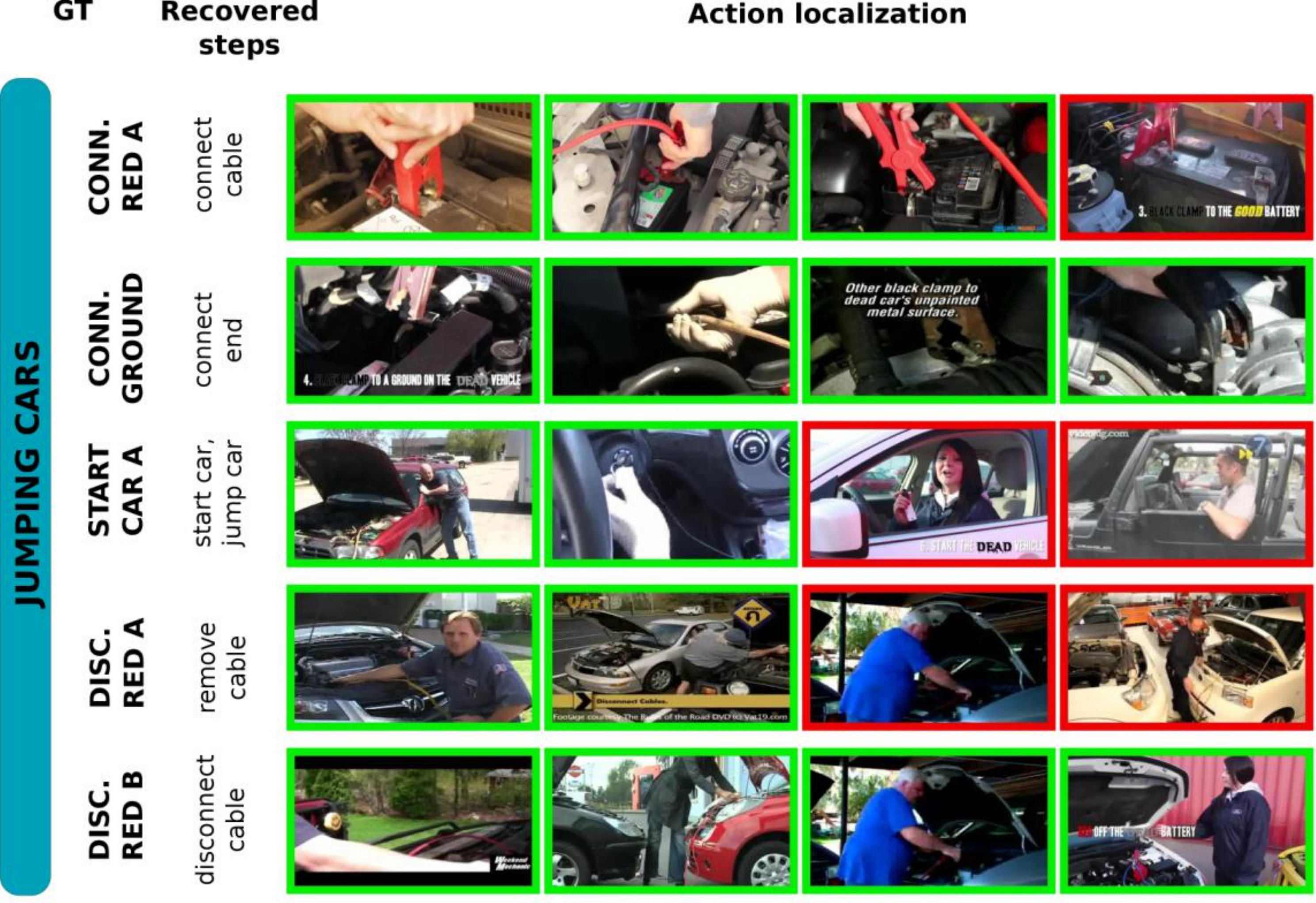}
\caption{\footnotesize {\bf Qualitative results for the task ``Jumping cars".}
}

\label{fig:qualitative_results_jump_car} 
\end{figure*}

\begin{figure*}[!ht]
\includegraphics[width=\linewidth]{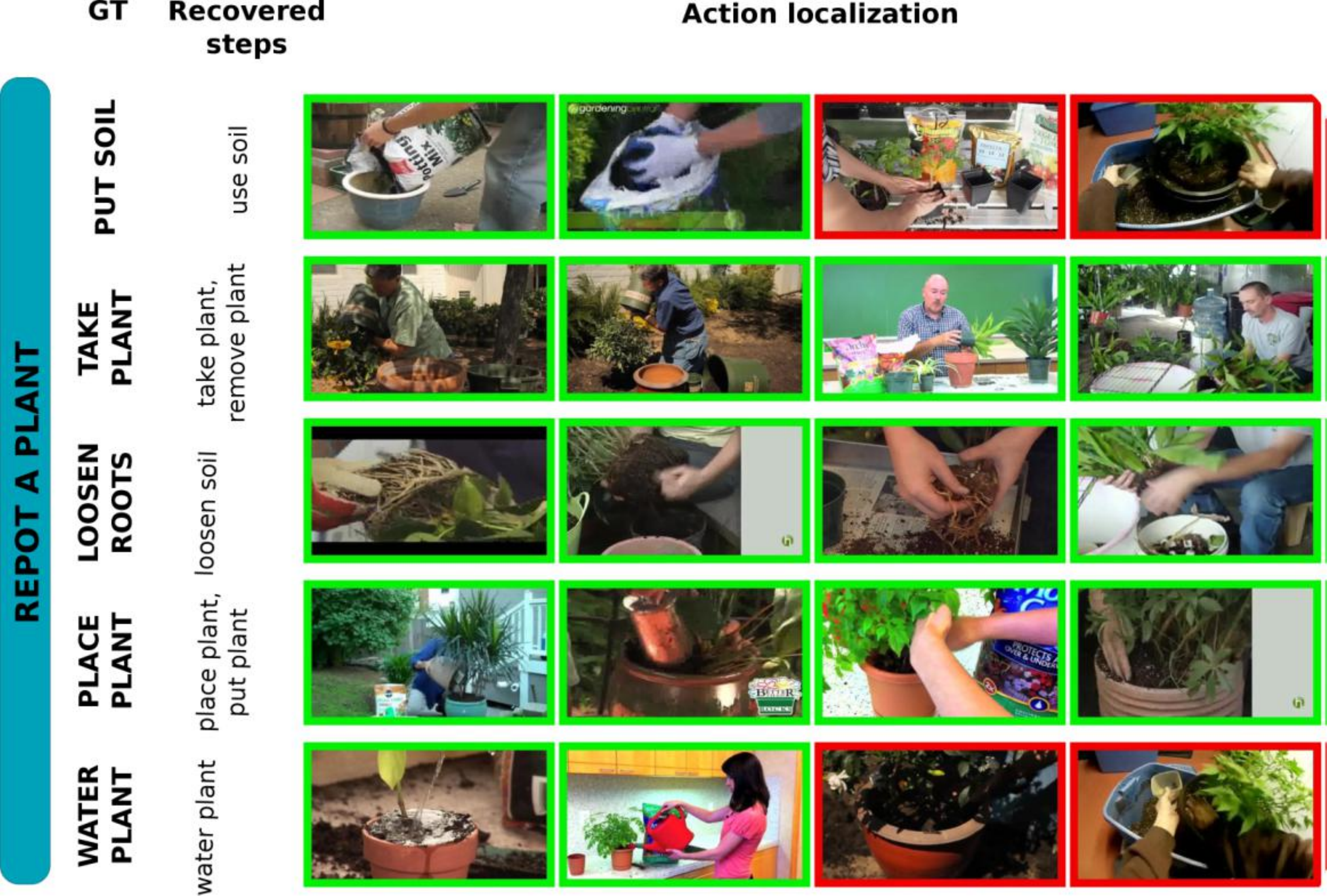}
\caption{\footnotesize {\bf Qualitative results for the task ``Repot a plant".}
}

\label{fig:qualitative_results_repot} 
\end{figure*}

\begin{figure*}[!ht]
\includegraphics[width=\linewidth]{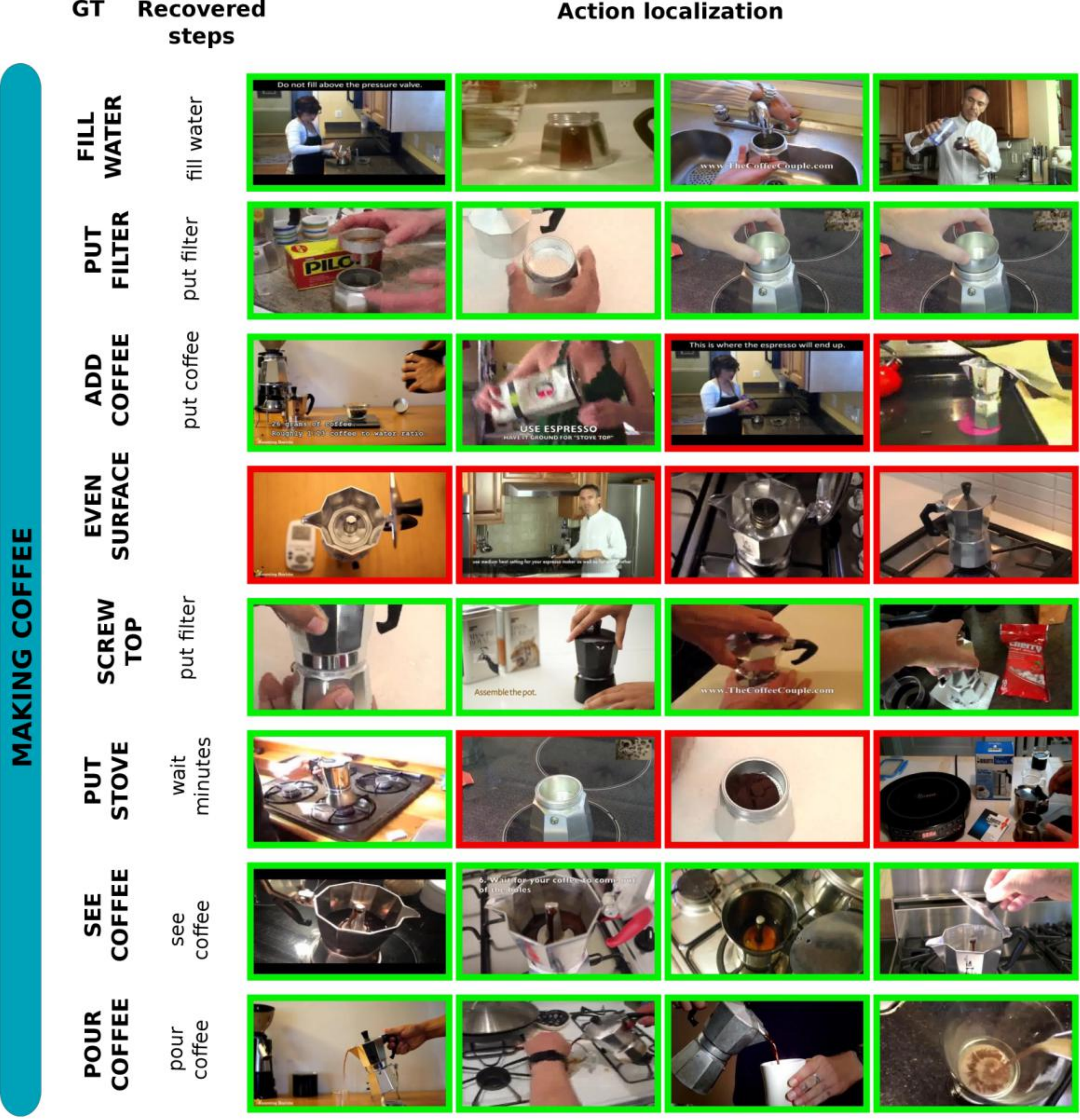}
\caption{\footnotesize {\bf Qualitative results for the task ``Making coffee".}
}
\label{fig:qualitative_results_coffee} 
\end{figure*}

\begin{figure*}[!ht]
\includegraphics[width=\linewidth]{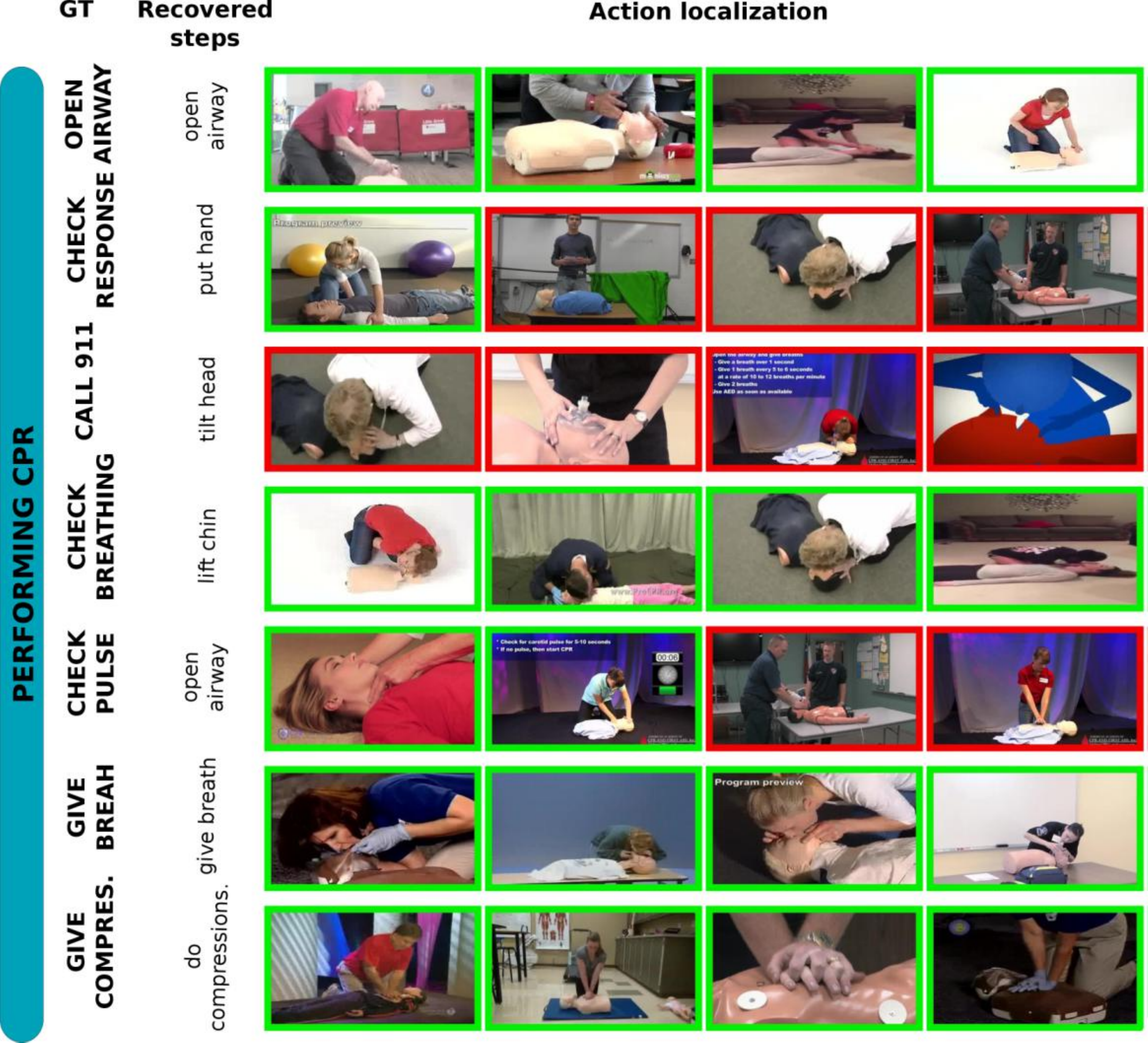}
\caption{\footnotesize {\bf Qualitative results for the task ``Performing CPR".}
}

\label{fig:qualitative_results_cpr} 
\end{figure*}

Examples of the recovered instruction steps for all five tasks are shown in Figure~\ref{fig:qualitative_results_changing_tire}--\ref{fig:qualitative_results_cpr}.
Each row shows one recovered step. 
For each step, we first show the clustered direct object relations, followed by representative example frames localizing the step in the videos. Correct localizations are shown in green. Some steps are incorrectly localized in some videos (red), but often look visually very similar.  Note how our method correctly recovers the main steps of the task and localizes them in the input videos.
Those results have been obtained by imposing $K\leq 10$ in our method.
The video on the project website illustrates action localization for the five tasks.

\bibliographystylesup{ieee}
\bibliographysup{biblio}

\newpage

\setlength{\tabcolsep}{2pt}

\begin{table*}
\footnotesize %

     \begin{subtable}{0.48\textwidth}
     \centering
\resizebox{0.95\textwidth}{!}{	
    \begin{tabular}{lrrrr >{\centering\hspace{0.5pt}}m{0cm}}
    \toprule
GT (\textbf{11}) & $K\leq 7$ &  $K\leq 10$ & $K\leq 12$ & $K\leq 15$\\
\cmidrule{1-5}
\textit{put brake on}  & &      &   &    \\ 
\textit{get tools out}  & & \textbf{get tire}     & \textbf{get tire}  & \textbf{get tire}   \\

\textit{start loose} & \textbf{loosen nut} & \textbf{loosen nut}     & \textbf{loosen nut}   &   \textbf{loosen nut}   \\ 

\textit{} &  &  & \textbf{} &  \textbf{lift car} \\

\textit{}          & \textbf{put jack} & \textbf{put jack}     &  \textbf{put jack} &   \textbf{put jack} \\  

\textit{}  &   &     &  \textbf{raise vehicle} &  \textbf{raise vehicle}\\  

\textit{jack car}  & \textbf{jack car}  & \textbf{jack car}     &  \textbf{jack car} &  \textbf{jack car}\\

\textit{unscrew wheel} & \textbf{remove nut}   & \textbf{remove nut}     &  \textbf{remove nut}  & \textbf{remove nut} \\

\textit{remove wheel} & & \textbf{take wheel}     &  \textbf{take wheel} & \textbf{take wheel} \\

\textit{put wheel}  & \textbf{take tire}  & \textbf{take tire}     & \textbf{take tire} &  \textbf{take tire} \\

\textit{screw wheel} & & \textbf{put nut}     & \textbf{put nut} &    \textbf{put nut} \\

\textit{lower car}  &  \textbf{lower jack} & \textbf{lower jack}     & \textbf{lower jack} &   \textbf{lower jack} \\  

\textit{}  &  & & & \textbf{remove jack}\\  

\textit{tight wheel}  & \textbf{tighten nut}  & \textbf{tighten nut}     &  \textbf{tighten nut}   &    \textbf{tighten nut}   \\  
\textit{put things back} &  &  & \textbf{take tire} &  \textbf{take tire} \\
\midrule
Precision  & 0.85  &  0.9    & 0.83 & 0.71   \\
Recall     & 0.54  &  0.9    & 0.9  & 0.9   \\

        \bottomrule
    \end{tabular}
}
     \caption{Changing a tire}
     \end{subtable}%
     \hspace*{\fill}%
	      \begin{subtable}{0.48\textwidth}
     \centering
\resizebox{0.95\textwidth}{!}{
    \begin{tabular}{lrrrr >{\centering\hspace{0.5pt}}m{0cm}}
    \toprule
GT (\textbf{10}) & $K\leq 7$ &  $K\leq 10$ & $K\leq 12$ & $K\leq 15$\\
\cmidrule{1-5}
\textit{grind coffee} & \textbf{•} & \textbf{•} & \textbf{•} & \textbf{•} \\
\textit{put filter} & \textbf{•} & \textbf{•} & \textbf{•} & \textbf{•} \\
\textit{add coffee} & \textbf{} & \textbf{put coffee} & \textbf{put coffee} & \textbf{put coffee} \\
\textit{even surface} & \textbf{•} & \textbf{•} & \textbf{•} & \textbf{•} \\
\textit{} & \textbf{} & \textbf{fill chamber} & \textbf{fill chamber} & \textbf{fill chamber} \\
\textit{} & \textbf{} & \textbf{} & \textbf{} & \textbf{make noise} \\
\textit{fill water} & \textbf{fill water} & \textbf{fill water} & \textbf{fill water} & \textbf{fill water} \\
\textit{screw top} & \textbf{} & \textbf{put filter} & \textbf{put filter} & \textbf{put filter} \\
\textit{} & \textbf{} & \textbf{} & \textbf{} & \textbf{fill basket} \\
\textit{} & \textbf{} & \textbf{see steam} & \textbf{see steam} & \textbf{see steam} \\
\textit{put stove} & \textbf{take minutes} & \textbf{take minutes} & \textbf{take minutes} & \textbf{take minutes} \\
\textit{•} & \textbf{make coffee} & \textbf{make coffee} & \textbf{make coffee} & \textbf{make coffee} \\
\textit{see coffee} & \textbf{see coffee} & \textbf{see coffee} & \textbf{see coffee} & \textbf{see coffee} \\
\textit{withdraw stove} & \textbf{•} & \textbf{•} & \textbf{•} & \textbf{turn heat} \\
\textit{pour coffee} & \textbf{make cup} & \textbf{make cup} & \textbf{make cup} & \textbf{make cup} \\
\textit{} & \textbf{•} & \textbf{•} & \textbf{•} & \textbf{pour coffee} \\
\midrule
Precision  & 0.8 &  0.67   & 0.67 &    0.54\\
Recall     & 0.4  &  0.6 & 0.6  &   0.7\\

        \bottomrule
    \end{tabular}
}
     \caption{Making coffee}
     \end{subtable}%
\vspace{3mm}
\hspace*{\fill}
\footnotesize %
 \begin{subtable}{0.48\textwidth}
     \centering
\resizebox{0.95\textwidth}{!}{
    \begin{tabular}{lrrrr >{\centering\hspace{0.5pt}}m{0cm}}
    \toprule
GT (\textbf{7}) & $K\leq 7$ &  $K\leq 10$ & $K\leq 12$ & $K\leq 15$\\
\cmidrule{1-5}
\textit{cover hole}      & \textbf{•}&  \textbf{•}   & \textbf{•}  & \textbf{take piece}   \\ 
    & \textbf{•}&  \textbf{•}   & \textbf{•}  & \textbf{keep soil}   \\ 
        & \textbf{•}&  \textbf{•}   & \textbf{•}  & \textbf{stop soil}   \\ 
        
\textit{take plant}   & \textbf{take plant}&  \textbf{take plant}  & \textbf{take plant}  & \textbf{take plant}  \\ 
\textit{put soil}     & \textbf{use soil}& \textbf{use soil}&\textbf{use soil}&\textbf{use soil}\\
\textit{loosen root}  & \textbf{loosen soil}&  \textbf{loosen soil}&\textbf{loosen soil}&\textbf{loosen soil}\\
\textit{place plant}  & \textbf{place plant}&   \textbf{place plant}&  \textbf{place plant}&  \textbf{place plant}\\
\textit{add top}      & \textbf{add soil}& \textbf{add soil}& \textbf{add soil}& \textbf{add soil}\\
& \textbf{•}&  \textbf{•}   & \textbf{•}  & \textbf{fill pot}   \\
& \textbf{•}&  \textbf{•}   & \textbf{•}  & \textbf{get soil}   \\ 
& \textbf{•}&  \textbf{•}   & \textbf{•}  & \textbf{give drink}   \\ 
\textit{water plant}  & \textbf{water plant}&  \textbf{water plant}&\textbf{water plant}&\textbf{water plant}\\
& \textbf{•}&  \textbf{•}   & \textbf{•}  & \textbf{give watering}   \\ 
\midrule
Precision  & 1    &  1      & 1    &  0.54  \\
Recall     & 0.86 &  0.86   & 0.86 &  1 \\

        \bottomrule
    \end{tabular}
}
     \caption{Repot a plant}
     \end{subtable}        
     \hspace*{\fill}%
     \begin{subtable}{0.48\textwidth}
          \centering
     \resizebox{0.95\textwidth}{!}{
      \footnotesize
    \begin{tabular}{lrrrr >{\centering\hspace{0.5pt}}m{0cm}}
    \toprule
GT (\textbf{7}) & $K\leq 7$ &  $K\leq 10$ & $K\leq 12$ & $K\leq 15$\\
\cmidrule{1-5}
\textit{open airway}  & \textbf{open airway}&    \textbf{open airway} & \textbf{open airway} & \textbf{open airway}   \\ 
\textit{check response}  & \textbf{•}&    \textbf{•}  &  \textbf{•} & \textbf{•}   \\ 
\textit{call 911}  & \textbf{•}&    \textbf{•}  &  \textbf{•} & \textbf{•}   \\ 
\textit{check breathing}  & \textbf{•}&    \textbf{•}  &  \textbf{•} & \textbf{•}   \\ 
\textit{check pulse}  & \textbf{}&    \textbf{put hand}  &  \textbf{put hand}  & \textbf{put hand}   \\ 
\textit{•}  & \textbf{tilt head} &    \textbf{tilt head}  &  \textbf{tilt head} & \textbf{tilt head}   \\  
\textit{}  &  \textbf{lift chin} &    \textbf{lift chin}  &  \textbf{lift chin} & \textbf{lift chin}   \\  
\textit{give breath}  & \textbf{give breath}&    \textbf{give breath}  &  \textbf{give breath}& \textbf{give breath}   \\ 
\textit{give compression}  & \textbf{do compr.}&    \textbf{do compr.}  &  \textbf{do compr.} & \textbf{do compr.}   \\ 
\textit{}  & \textbf{open airway}&    \textbf{open airway}  &  \textbf{open airway} & \textbf{open airway}   \\  
\textit{}  & \textbf{•}&    \textbf{start compr.}  &  \textbf{start compr.} & \textbf{start compr.}   \\  
\textit{}  & \textbf{•}&    \textbf{}  &  \textbf{} & \textbf{continue cpr}   \\ 
\textit{}  & \textbf{•}&    \textbf{do compr.}  &  \textbf{do compr.} & \textbf{do compr.}   \\ 
\textit{}  & \textbf{•}&    \textbf{•}  &  \textbf{} & \textbf{put hand}   \\ 
\textit{}  & \textbf{•}&    \textbf{give breath}  &  \textbf{give breath} & \textbf{give breath}   \\ 

\midrule
Precision  &  0.	5 &  0.4      &  0.4 &  0.33  \\
Recall     &  0.43 &  0.57    & 0.57 & 0.57   \\
\bottomrule
    \end{tabular}
   } 
     \caption{Performing CPR}
     \end{subtable}

\centering
\footnotesize %

     \begin{subtable}{0.48\textwidth}
     \centering
\resizebox{0.95\textwidth}{!}{     
    \begin{tabular}{lrrrr >{\centering\hspace{0.5pt}}m{0cm}}
    \toprule
GT (\textbf{12}) & $K\leq 7$ &  $K\leq 10$ & $K\leq 12$ & $K\leq 15$\\
\cmidrule{1-5}
\textit{get cars}  & \textbf{}&    \textbf{}  &  \textbf{} & \textbf{}   \\ 
\textit{open hood}  & \textbf{}&    \textbf{}  &  \textbf{} & \textbf{}   \\ 
\textit{}  & \textbf{}&    \textbf{}  &  \textbf{} & \textbf{have terminal}   \\  
\textit{}  & \textbf{}&    \textbf{}  &  \textbf{attach cab.} & \textbf{attach cab.}   \\ 
\textit{connect red A}  & \textbf{connect cable}&    \textbf{conn. cable}  &  \textbf{conn. cable} & \textbf{conn. cable}   \\
\textit{}  & \textbf{}&    \textbf{}  &  \textbf{} & \textbf{conn. clamp}   \\  
\textit{}  & \textbf{charge battery}&    \textbf{charge batt.}  &  \textbf{charge batt.} & \textbf{charge batt.}   \\ 
\textit{connect red B}  & \textbf{connect end}&    \textbf{conn. end}  &  \textbf{conn. end} & \textbf{conn. end}   \\ 
\textit{connect black A}  & \textbf{}&    \textbf{}  &  \textbf{conn. cab.} & \textbf{conn. cab.}   \\ 
\textit{connect ground}  & \textbf{•}&    \textbf{•}  &  \textbf{have cab.} & \textbf{have cab.}   \\ 
\textit{start car A}  & \textbf{start car}&    \textbf{start car}  &  \textbf{start car} & \textbf{start car}   \\ 
\textit{start car B}  & \textbf{•}&    \textbf{•}  &  \textbf{start vehicle} & \textbf{start veh.}   \\ 
\textit{}  & \textbf{•}&    \textbf{•}  &  \textbf{start engine} & \textbf{start eng.}   \\ 
\textit{remove ground}  & \textbf{remove cable}&    \textbf{rem. cable}  &  \textbf{rem. cable} & \textbf{rem. cable}   \\ 
\textit{remove black A}  & \textbf{disconnect cable}&    \textbf{disc. cable}  &  \textbf{disc. cable} & \textbf{disc. cable}   \\ 
\textit{remove red B}  & \textbf{}&    \textbf{}  &  \textbf{} & \textbf{}   \\ 
\textit{remove red A}  & \textbf{}&    \textbf{}  &  \textbf{} & \textbf{}   \\ 
        
        \midrule
Precision  & 0.83  &  0.83   &  0.72 &  0.69  \\
Recall     & 0.42  &  0.42   & 0.67 &  0.67 \\
\bottomrule
    \end{tabular}
}
     \caption{Jumping cars}
     \end{subtable}%

 \caption{\small
  Automatically recovered sequences of steps for the five tasks considered in this work. 
        Each recovered step is represented by one of the aligned direct object relations  (shown in bold). 
        Note that most of the recovered steps correspond well to the ground truth steps (showed in italic).
        The results are shown for setting the maximum number of discovered steps, $K = \{7,10,12,15\}$. 
        Note how our method automatically selects less than $K$ steps in some cases.
         These are the automatically chosen $k\leq K$ steps that are the most salient in the aligned narrations as described in Sec.~\ref{subsec:model_text}.  
        }    
    \label{tab:script_res} 
\end{table*}

\setlength{\tabcolsep}{6pt}

\end{document}